%% file: 0_excavation_RL.tex
\documentclass[letterpaper, 10 pt, journal, twoside]{IEEEtran}
\usepackage{amsmath}
\usepackage{amssymb}
\usepackage{bm}
\usepackage{graphicx}
\usepackage{subcaption}
\usepackage[skip=0pt,font=small,labelfont=bf]{caption}
\usepackage{wrapfig}
\usepackage{mwe}
\usepackage{todonotes}
\usepackage{adjustbox}
\usepackage{pgfplots}
\usetikzlibrary{patterns}
\usepackage{color,soul}
\usepackage{cellspace}
\usepackage{amssymb}
\usepackage{url}
\usepackage{breqn}
\usepackage{multirow}
\usepackage[top=72pt, left=54pt, right=54pt, bottom=54pt]{geometry}
\usepackage{cite} 
\usepackage[ruled,linesnumbered]{algorithm2e}
\usepackage[noend]{algpseudocode}
\usepackage{booktabs}
\usepackage{soul}

\usepackage{fixme}
\fxsetup{status=draft, theme=color}
\definecolor{fxtarget}{rgb}{0.8000,0.0000,0.0000}
\definecolor{fxnote}{rgb}{0.0000,0.0000,0.8000}

\begin{document}

\title{
	Excavation Reinforcement Learning Using Geometric Representation
}

\author{Qingkai Lu$^*$$^{1}$, Yifan Zhu$^*$$^{1,2}$, Liangjun Zhang$^{1}$
\thanks{*These authors contributed equally.}
\thanks{Manuscript received: September, 9, 2021; Revised December, 12, 2021; Accepted January, 13, 2022.}
\thanks{This paper was recommended for publication by Editor H. Liu upon evaluation of the Associate Editor and Reviewers' comments.
This work was supported by Baidu Research USA.} 
\thanks{$^{1}$Q. Lu, Y. Zhu, and L. Zhang are
  with the Robotics and Auto-Driving Lab, Baidu Research, Sunnyvale, CA USA. 
 {\tt\footnotesize \{qingkailu, yifanzhu, liangjunzhang\}@baidu.com.}}%
\thanks{$^{2}$: Y. Zhu is also with the Departments of Computer Science, University of Illinois at Urbana-Champaign, IL, USA.
        {\tt\small yifan16@illinois.edu}}%
\thanks{Digital Object Identifier (DOI): see top of this page.}
}

\markboth{IEEE Robotics and Automation Letters. Preprint Version. Accepted January, 2022}
{Lu \& Zhu \MakeLowercase{\textit{et al.}}: Excavation Reinforcement Learning Using Geometric Representation} 

\maketitle


\begin{abstract}
Excavation of irregular rigid objects in clutter, such as fragmented rocks and wood blocks, is very challenging due to their complex interaction dynamics and highly variable geometries. In this paper, we adopt reinforcement learning (RL) to tackle this challenge and learn policies to plan for a sequence of excavation trajectories for irregular rigid objects, given point clouds of excavation scenes.
Moreover, we separately learn a compact representation of the point cloud on geometric tasks that do not require human labeling. We show that using the representation reduces training time for RL, while achieving similar asymptotic performance compare to an end-to-end RL algorithm. When using a policy trained in simulation directly on a real scene, we show that the policy trained with the representation outperforms end-to-end RL. To our best knowledge, this paper presents the first application of RL to plan a sequence of excavation trajectories of irregular rigid objects in clutter.
\end{abstract}




\section{Introduction}
\label{sec:intro}
\input{1_introduction}

\section{Related Work}
\label{sec:related_work}
\input{2_related_work}
\section{Problem Definition}
\label{sec:problem_define}
\input{3_problem_definition}



\section{Reinforcement Learning for Excavation}
\label{sec:geometry_representaion}
\input{4_geometric_representation}


\section{Excavation Experiments Setup}
\label{sec:exp_setup}
\input{6_exp_setup}

\section{Excavation Experimental Results}
\label{sec:exp_results}
\input{7_exp_res}

\section{Conclusion}
\label{sec:discussion}

\input{8_discussion}

%



\bibliographystyle{IEEEtran}
\bibliography{references}  


\clearpage
\section{Appendix}
\label{sec:appendix}
\input{9_appendix}

\clearpage

\end{document}

%% file: 1_introduction.tex
\begin{figure}[t!]
    \centering
    \includegraphics[trim=0cm 2cm 0cm 1cm,clip,width=0.4\textwidth]{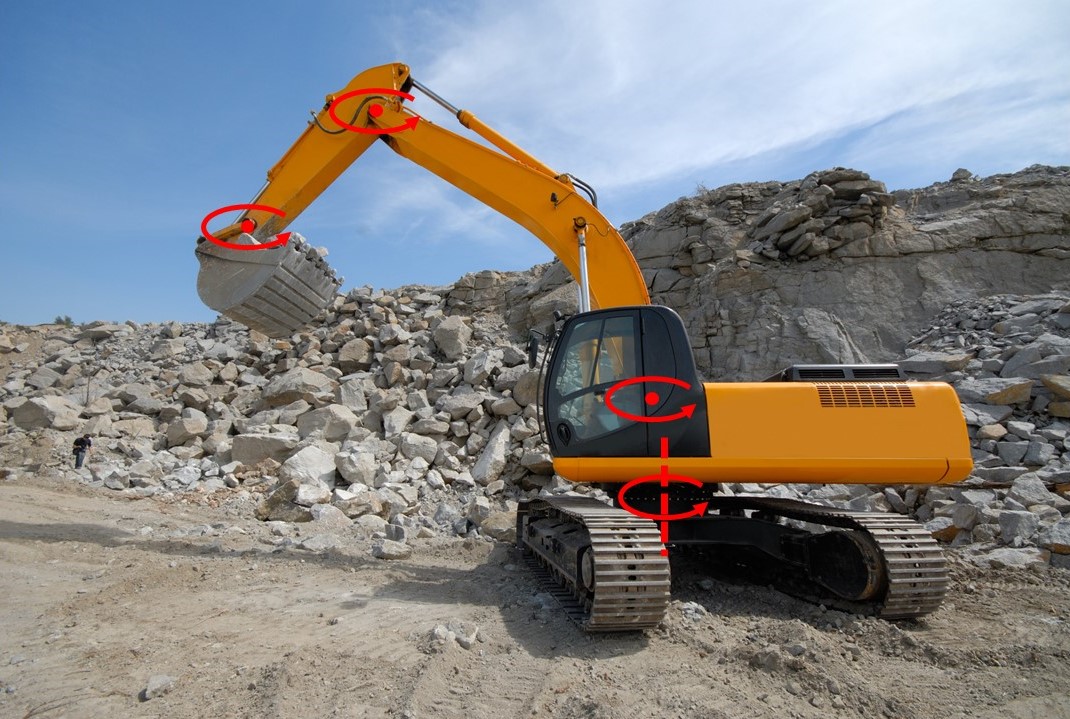}
    \vspace{10px}
    \caption{The figure shows an excavator excavating fragmented rocks, with joints labeled in red. An excavator has a total of 4 degrees of freedom, excluding the mobile base.}
    \label{fig:excavator_rock}
\end{figure}

\IEEEPARstart{A}{s} an integral part of construction, mining, and various other important engineering fields, excavator operation requires skilled workers and often needs to be operated in extreme outdoor conditions that lead to injuries~\cite{Marsh2015}. There have been an increasing amount of research efforts in excavation automation~\cite{Zhang2021, egli2020towards, sandzimier2020data}. However, the majority of the literature only considers the excavation of homogeneous granular materials such as soil and sand, and the excavation of irregular rigid objects such as fragmented rocks is not as explored. An example of an excavator excavating fragmented rocks is shown in Fig.~\ref{fig:excavator_rock}.

A major challenge in excavation, especially for materials such as fragmented rocks, is that the terrain-machine physical interaction is extremely hard to model~\cite{dadhich2016key}. This is different from excavation for homogeneous granular materials such as sand, whose interaction physics with an excavator bucket can be modeled or learned to a certain extent~\cite{Cannon2000,Tan2005}. Moreover, cluttered rigid objects have highly variable geometric shapes and only the objects on the surface can be partially observed, which makes the perception for excavation challenging. 
In order to address these challenges, learning-based\cite{Lu2021Excavation, sotiropoulos2020autonomous} and admittance-control-based\cite{dobson2017admittance,fernando2019iterative} approaches have been proposed for rigid objects excavation. However, the existing excavation learning work only focus on maximizing the excavation of a single excavation. 
Excavations are usually performed in sequence for given excavation tasks such as material loading and trenching. Moreover, the planning and execution of one excavation depends on the current excavation scene and affects future excavations. 
In comparison, we model sequential excavation planning as a Markov decision process (MDP) that considers the long term excavation reward, and use reinforcement learning (RL) to plan for a series of digging trajectories in this work. 

This paper mainly has several contributions. First, we use RL to learn to sequentially plan excavation trajectories for rigid objects in clutter. Compared to existing works that adopt RL to plan controls to execute a single trajectory\cite{Azulay2021,Dadhich2016,Kurinov2020}, our paper presents the first RL work for a sequence of excavation trajectories of cluttered rigid objects. Second, we learn a novel geometric representation for point clouds without human labeling, which allows us to use a small policy network and reduce RL training time, while achieving similar asymptotic performance as compared to an end-to-end RL algorithm that takes the raw point cloud as input. Finally, we perform excavation experiments for cluttered objects in simulation and the real world, and demonstrate that the learned representation trained only in simulation transfers better than end-to-end RL to the physical experiments. 


The literature is reviewed in Section~\ref{sec:related_work}. We introduce the formal problem definition for cluttered rigid objects excavation in Section~\ref{sec:problem_define}. We then give a thorough account of RL with geometric representation learning in Section~\ref{sec:geometry_representaion}. We present the excavation experiments setup and results in Section~\ref{sec:exp_setup} and~\ref{sec:exp_results} respectively. Finally, we make a brief conclusion in Section~\ref{sec:discussion}. 

%% file: 2_related_work.tex
\textbf{Excavation Planning and Learning:}
The majority of prior work on excavation automation explores the excavation of homogeneous materials such as soil and sand. Here we review literature on soil and sand excavation first, which could be roughly grouped into 1) control of hydraulic earthmoving machines~\cite{egli2020towards,chang2002straight, maeda2015combined,sotiropoulos2019model}; 2) trajectory-level planning~\cite{yang2021optimization,sandzimier2020data}; 3) task-level planning and excavation system engineering~\cite{stentz1999robotic,Jud2017,Seo2015TaskPD,Zhang2021}. There have have been a relatively small amount of works that focus on rigid objects excavation, partly because of the complex, stochastic nature of contact dynamics. Dobson et al. controls earthmoving machines for fragmented rock excavation with admittance control~\cite{dobson2017admittance}. The follow-up work improves the algorithm by adapting the parameters used in admittance control online with iterative learning~\cite{fernando2019iterative}. 
In~\cite{sotiropoulos2020autonomous}, the authors explore the challenge of excavating a single rock partially embedding in sand using Gaussian process regression and an unscented Kalman Filter.
Lu and Zhang tackle the problem of excavation of rigid objects in clutter with a learning based approach where a excavation success prediction model is learned from data, which is used in the cross-entropy method for trajectory planning~\cite{Lu2021Excavation}. Dadhich et al. pose a imitation learning-based approach for learning expert demonstrations of rock excavation with wheel loaders~\cite{Dadhich2016}. Different from these approaches which focus on the control or planning of a single excavation, our approach models a sequence of excavations with an MDP and use RL to plan a sequence of excavation trajectories.

\textbf{Learning Geometric Quantities}:
Learning the normal and/or curvature from RGB images is a common task in computer vision. Wang et al. propose to perform single RGB image normal estimation with CNN~\cite{Wang2015}. Eigen and Fergus also use CNN to perform joint depth, normal estimation and semantic segmentation of a single RGB images with CNN~\cite{Eigen2015}. Similarly Dharmasiri et al.~\cite{Dharmasiri2017} and Qi et al.~\cite{Qi2018} also adopt CNN to perform joint depth and normal estimation of single RGB images.

\textbf{}{Learning on Point Clouds}:
With the seminal work of PointNet by Qi et al.~\cite{Qi2017a}, a general feature extractor of point clouds, deep learning on point clouds have attracted significant research interests especially given increasingly cheap and reliable sensors that capture point clouds, and are reviewed in recent surveys~\cite{Liu2019,Lu2020,Guo2020}. Given the good performance across different tasks and benchmarks, and implementation simplicity of PointNet++~\cite{QiPointnet++}, in this work we adopt it to learn the representation for the task of excavation.

\textbf{Representation Learning for Manipulation Tasks}:
While it has been shown that one can learn a control policy end-to-end using deep reinforcement learning (DRL) given high-dimensional observations\cite{Wang2020DRL}, a significant, sometimes prohibitive amount of data is needed. However, it is possible to take advantage of compact, low-dimensional state representation to improve data efficiency~\cite{Munk2016}. Such state representation learning is usually performed by pre-training on different tasks~\cite{Lee2020} or introducing additional tasks by adding to the RL rewards~\cite{DeBruin2018}. A recent survey reviews the different types of additional learning objects and models of state representation learning~\cite{Lesort2018}. Common objectives include reconstruction, forward and inverse dynamics prediction, physical priors such as slowness, etc. Few work, however, has explored to adopt tasks of normal and curvature estimation and object counting. In addition, the majority of existing literature only considers RGB images as the visual observation while we consider point clouds. 

%% file: 3_problem_definition
In this work, we focus on planning a sequence of excavation trajectories for rigid objects in clutter. Given the visual representation of the excavation scene $Z$, in our case a point cloud, the goal is to plan N digging trajectories $T_1, \dots, T_{N}$ in task (Cartesian) space and maximize the accumulative volume of the excavated rigid objects. Both the scene $Z$ and trajectory $T$ are defined in the world frame, which are shown in Fig.~\ref{fig:scene_setup}.



\begin{figure}[h]
    \centering
    \includegraphics[width=0.45\textwidth]{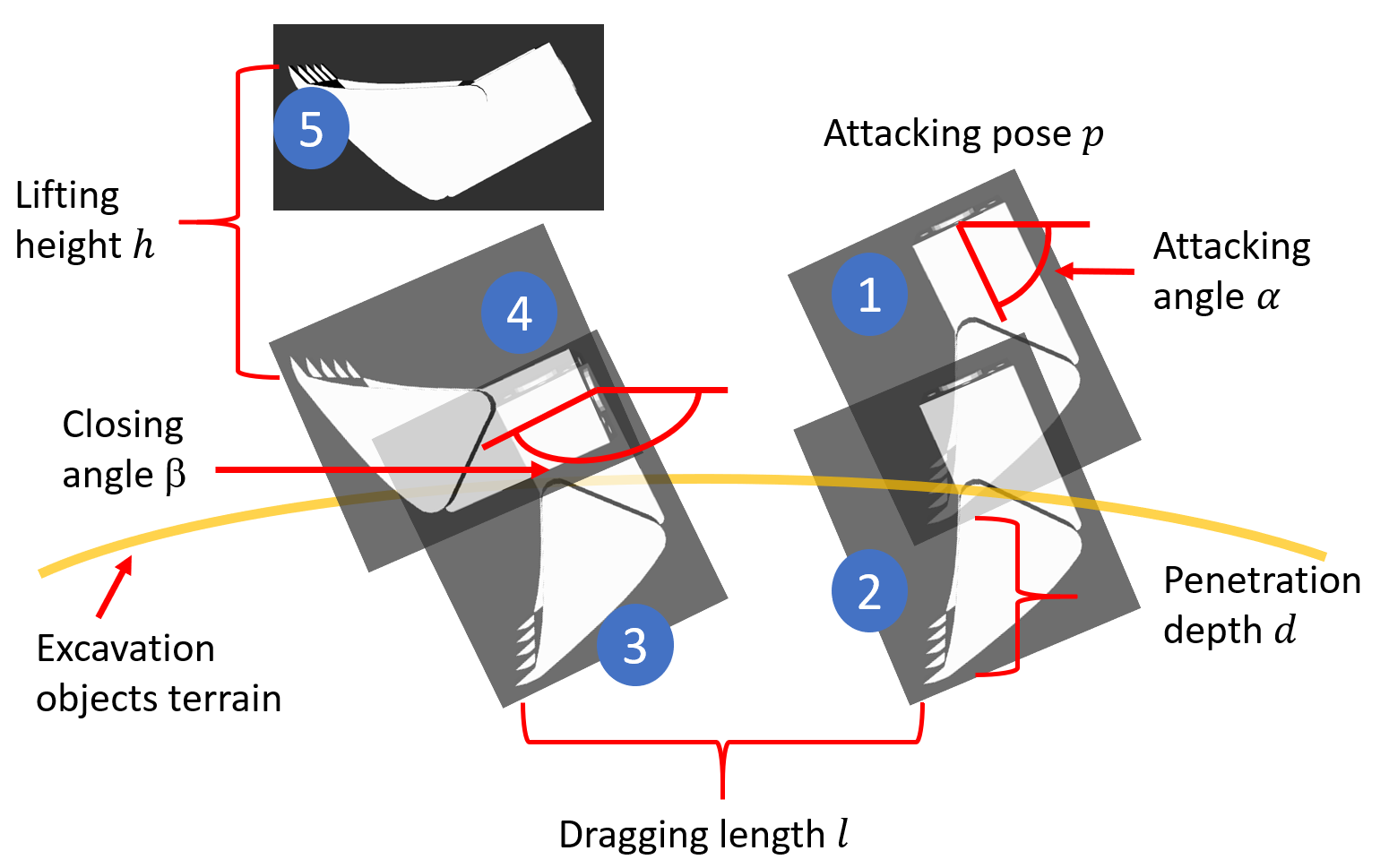}
    \caption{This figure adapted from~\cite{Lu2021Excavation} visualizes our excavation trajectory representation in the task space. The numbers in blue circles represent the sequence of our five excavation phases. The bucket poses with ID from $1$ to $5$ represents the attacking, penetration, dragging, closing, and lifting phase respectively.}
    \label{fig:exv_task_traj}
\end{figure}

We divide an excavation trajectory into 5 phases: attacking, digging, dragging, closing, and lifting, following common practice in the literature~\cite{Lu2021Excavation,sing1995synthesis}, shown in Fig.~\ref{fig:exv_task_traj}. The excavator base stays still during the entire trajectory and the excavator starts the attacking phase at an attacking pose, which can be fully specified by the position $(x, y, z)$ and the attack angle $\alpha$. During the attacking phase, the bucket penetrates the substrate in the gravity direction for a distance of $d$. In the dragging phase, the bucket moves a distance of $l$ towards the excavator base while fixing the orientation of the bucket. Finally, during the closing and lifting phases, the bucket is rotated to an angle of $\beta$, followed by lifting the bucket up a height of $h$. 

Since we always start the attacking phase on the surface objects clutter in this work, given the $(x, y)$ coordinates of the attacking pose $p$, we can get its $z$ coordinate from the depth image of the RGB-D sensor. In addition, we assume the excavator always plans a trajectory that drags as much as possible without inverse kinematics (IK) failure, self-collision or collisions with the object digging tray capped at a maximum dragging length $l_{max}$, which is empirically set and fixed for all excavation trajectories. Finally, we select a fixed closing angle and always lift to a fixed height w.r.t. the robot since we assume that the closing angle and lifting height has a small influence on the quality of excavation. As a result, we simplify the excavation trajectory $T$ as its attacking pose: $(x, y, \alpha)$, which we also refer to as the attacking pose.

%% file: 4_geometric_representation.tex
Excavation tasks such as material loading and trenching are sequential as the planning of one excavation depends on the current excavation scene and affects future excavations. Therefore, we naturally model excavation planning of attacking poses as a MDP problem and solve it using RL. 

While deep RL algorithms have demonstrated their capability to learning manipulation policies from raw observations, they usually require a large amount of data and computational resources~\cite{ibarz2021train}. To improve the training efficiency of excavation reinforcement learning, we propose to learn a low-dimensional representation of the observations by utilizing geometric supervision without laborious manual labeling. 

\subsection{Excavation Reinforcement Learning}
\label{sec:exv_rl}
We model excavation planning of attacking poses as a MDP problem $<S, A, R, P, \lambda>$, where $S, A, R, P,$ and $\lambda$ represent the state space, action space, reward, transition function, and discount factor. The point cloud of an excavation scene is defined to be its state $s \in S$. An action $a \in A$ executes a excavation trajectory $T$. The reward $r$ consists of the excavated objects volume. 

We use the proximal policy optimization (PPO) algorithm~\cite{schulman2017proximal} to learn to generate attacking poses for rigid objects excavation. As an actor-critic approach, PPO optimizes a surrogate objective function using stochastic gradient ascent. In each episode, we use a new excavation scene and allow the excavator to dig 10 times before starting a new episode and scene. 

\subsection{Geometric Representation Learning for Excavation}
\label{sec:geo_rep_rl}
To learn a compact representation that captures useful features for the excavation task,
we propose a deep network based on PointNet++\cite{QiPointnet++} that takes a point cloud as input and predicts its normal, curvature, and the number of objects in the digging tray. 
Excavation of cluttered rigid objects requires reasoning over both local and global features of the excavation scene. Globally, the robot needs to select the best excavation location to maximize the volume of excavated objects. 
At the same time, bucket-terrain interaction needs to consider the local object geometry. Therefore, we choose the task of point cloud normal and curvature prediction to encourage the capture of local features and task of counting the total number of objects in the scene to reason about the entire excavation scene. Note that although only the surface point cloud of the excavation scene is visible, the average size of the objects is similar between different scenes, which enables one to estimate the total number of objects. 

\begin{figure}[ht]
\centering
\includegraphics[trim=0.3cm 2cm 6cm 0.5cm,clip,width=.98\linewidth]{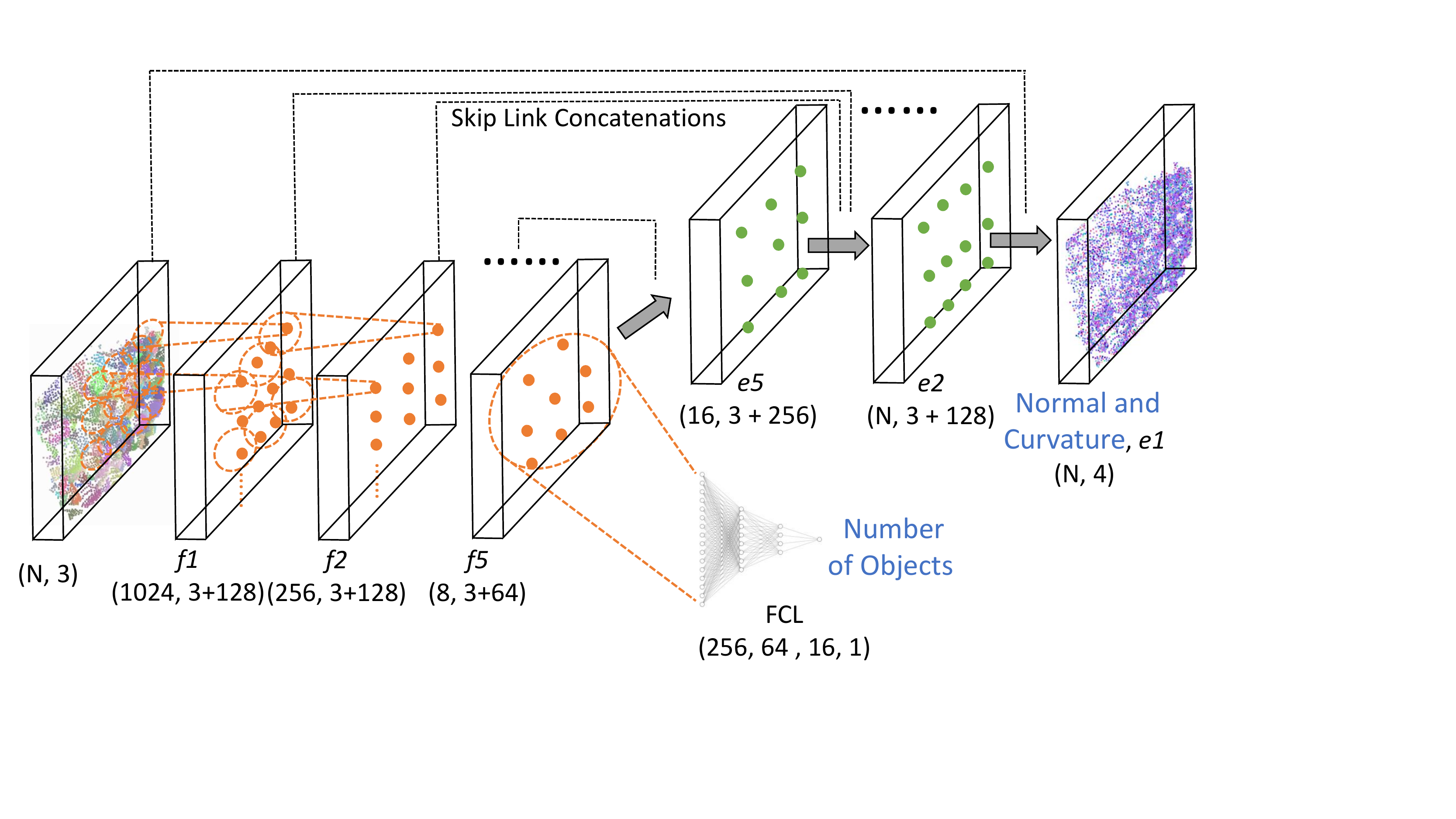}
\caption{\label{fig:representation_learning_architecture} Representation learning architecture.}
\end{figure}

There have been an increasing amount of literature on featuring learning of unstructured point clouds~\cite{Liu2019}, and we adopt PointNet++ for its simplicity and performance. The architecture of our representation network is shown in Fig.~\ref{fig:representation_learning_architecture}, which is a modified version of the original PointNet++. We encourage readers to refer to~\cite{QiPointnet++} for more details. We first use farthest point sampling (FPS) to downsample to $7,000$ points from the raw point cloud. The FPS point cloud is then fed into the representation network and successively sampled and grouped into a smaller number of points that, in addition to its 3D position, has an extra feature vector of certain dimensions, which are extracted with PointNet~\cite{Qi2017a}. In particular, there are a total of five feature extraction layers ($f1$ - $f5$) with 1024, 256, 64, 16, and 8 points, respectively. The corresponding numbers of features for each point of the five layers are 128, 128, 128, 64, and 32. Therefore, we encode the point cloud into 8 final points and each point has a representation feature vector of 35 (32D feature + 3D position) dimensions. We concatenate the points and use it as a 280-D representation. 

For normal and curvature predictions, the extracted features are successively propagated into the original set of points in the point cloud using the feature propagation layers ($e5$ - $e1$), and the number of points in layer $ei = fi$. In each layer, the feature vector at each point is achieved by interpolating the point features in the previous layer to get a feature vector at each points in the current layer, concatenating with skip linked point features from the set abstraction level, and extracting a final feature with unit PointNet. The dimension of point feature in each feature propagation layers are 256, 256, 256, 128, and 4 (4 for normal and curvature predictions). For prediction of the total number of objects in the digging scene, the extracted features of the 8 points are first passed through a PointNet then several fully-connected layers (FCL). 

All feature extraction and propagation layers use $tanh$ as the activation function, while all the other layers use ReLU. The total loss is a weighted sum of normal, curvature and object counting losses, with $\mathcal{L}_{total} = w\cdot\mathcal{L}_{normal} + \mathcal{L}_{curv} + \mathcal{L}_{obj}$. Both $\mathcal{L}_{curv}$ and $\mathcal{L}_{obj}$ are smooth L1 losses and we set $w$ = 10. To compute the normal loss, we normalize the raw normal prediction and take the negative of the dot product with the ground truth.
We use an analytical method to compute the principal curvature and normal ground truth~\cite{RusuDoctoralDissertation}. Moreover, the number of objects can be counted automatically in simulation. As a result, we can train the representation model without human labeling. 

To integrate this learned representation to a RL algorithm, we use the feature extraction layers as the feature extractor, and keep the weights fixed during RL learning, where only the policy network weights get updated.

%% file: 6_exp_setup.tex
\begin{figure}[h]
    \centering
    \subfloat[\centering Excavation Setup]{\includegraphics[trim=0cm 0cm 0cm 0cm,clip,width=.4825\linewidth]{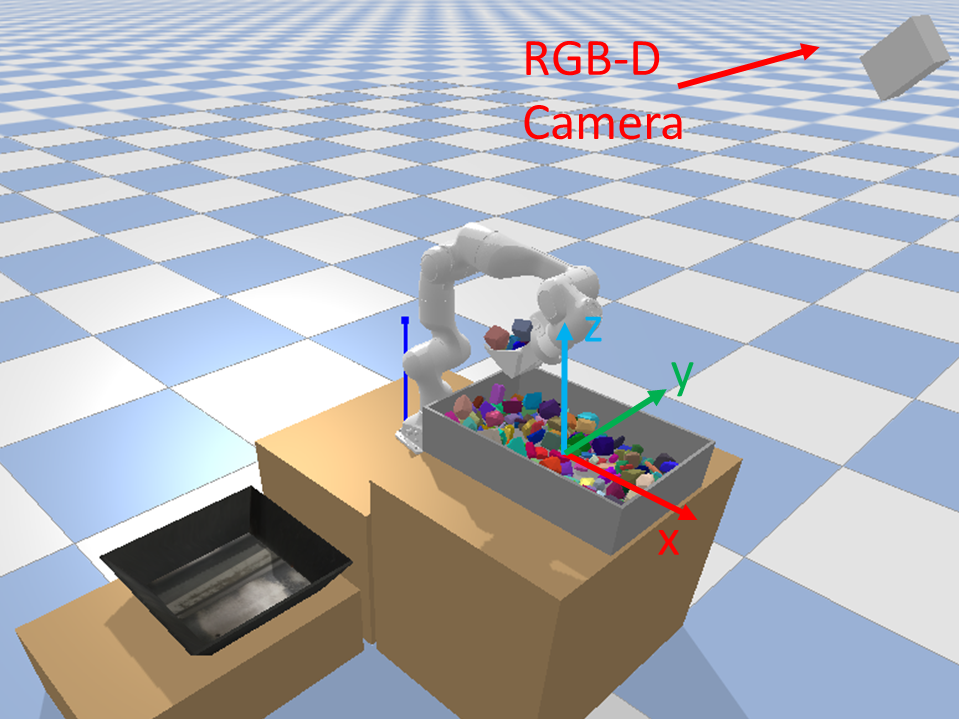}\label{fig:sim_scene}}
    \subfloat[\centering Digging Trajectory]{\includegraphics[trim=0cm 0cm 0cm 0.cm,clip,width=.48\linewidth]{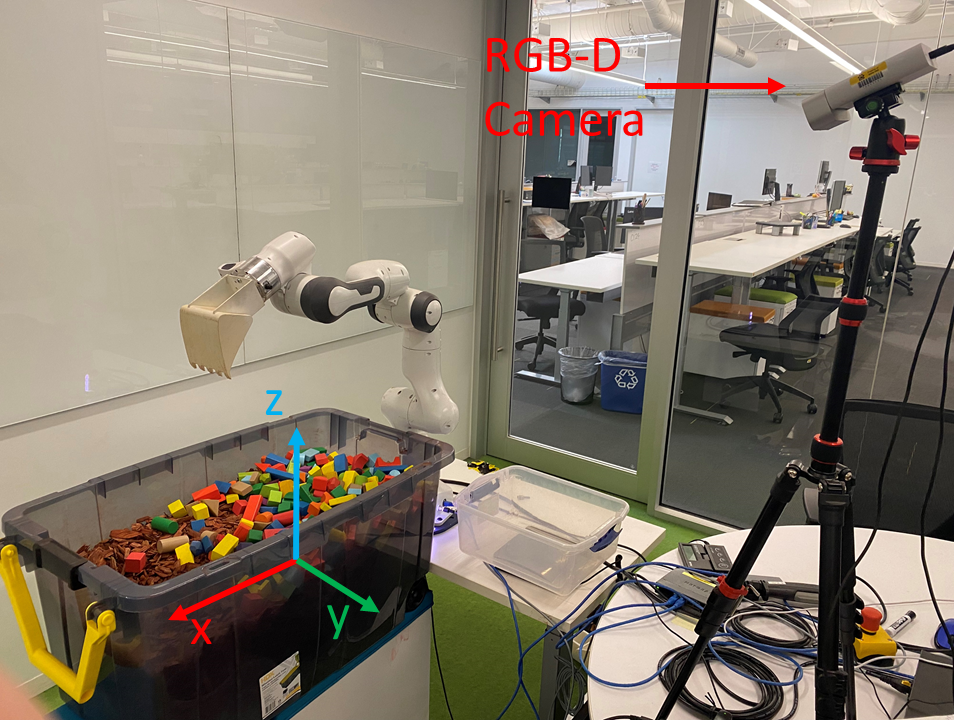}\label{fig:real_scene}}
    \caption{The experiment scenes in simulation and the real world. We plot the tray coordinate for simulation and the real world.}
    \label{fig:scene_setup}
\end{figure}

Throughout the experiments, we emulate an excavator with a 7-degrees-of-freedom (DoF) Franka Panda robot arm mounted with a digging bucket, shown in Fig.~\ref{fig:scene_setup}. We only use the shoulder panning, shoulder lifting, elbow lifting , and the wrist lifting joints. We fixate the joint positions of the other 3 joints. We set up both simulation and physical experiments, shown in Fig.~\ref{fig:scene_setup}. 
The details of the experiments setup are described in Section~\ref{sec:appendix} of the Appendix.

\vspace{-5pt}
\vspace{-5pt}

\subsection{Excavation Workflow}\label{sec:workflow}
We conduct both simulation and physical experiments using the same workflow. At the start of the scene, the robot moves to a home configuration and the RGB-D camera captures of a point cloud of the rigid objects without visual occlusion from the robot arm. The point cloud is cropped to the range of excavation tray, and subsequently converted to a grid/height map such that we can obtain the height $z$ of a 2D attacking point $(x,y)$. In all of our experiments we downsample the point cloud to 7,000 points using FPS before feeding into the PointNet++-based feature extractor. The policy network of RL then plans an attacking pose given the current scene, after which a robot joint trajectory $\Tilde{T}$ is calculated and sent to the robot controller. Both the point cloud and the attacking pose $(x, y, \alpha)$ are represented in the digging tray frame. 

Given a trajectory $T$ in task space, we compute a discretized open-loop robot joint trajectory that will be sent to the robot, defined as $\Tilde{T} = \{(q_0, t_0), \dots, (q_{k-1}, t_{k-1})\}$, which is $k$ pairs of robot joint configuration $q\in \mathbb{R}^7$ and time stamps.  The joint trajectory $\Tilde{T}$ is computed first by calculating the discretized task space trajectory based on $T$, where the bucket follows constant desired translation and rotation speeds, and the control rate is 100\,Hz. Then we solve for each robot joint configuration $q$ using inverse kinematics (IK). A planning failure occurs when an IK solution could not be found or when there is self-collision or collision between the robot and the objects digging tray. The trajectory is sent to the robot controller in an open-loop fashion where we also take advantage of the internal joint-space impedance controller of the Panda robot with joint stiffness set at $100$\,Nm/rad.

%% file: 7_exp_res.tex
In this section, we first report the experiments of the representation learning. Then we describe the RL training details and results. Finally, the excavation experiments in simulation and the real world are discussed.
We have released the trained RL excavation policies, experiment data, and experiment videos~\footnote{https://drive.google.com/drive/folders/19n-V573He55i6WqnoINukvoaU1oqSUmQ?usp=sharing}.

\subsection{Representation Learning}
\begin{figure*}[]
\vspace{-10px}
\centering
\setlength{\tabcolsep}{0px}
\begin{tabular}{lllllc}
{\raisebox{0.0\height}{\includegraphics[trim=0cm 0cm 0cm 0cm,clip,width=.18\linewidth]{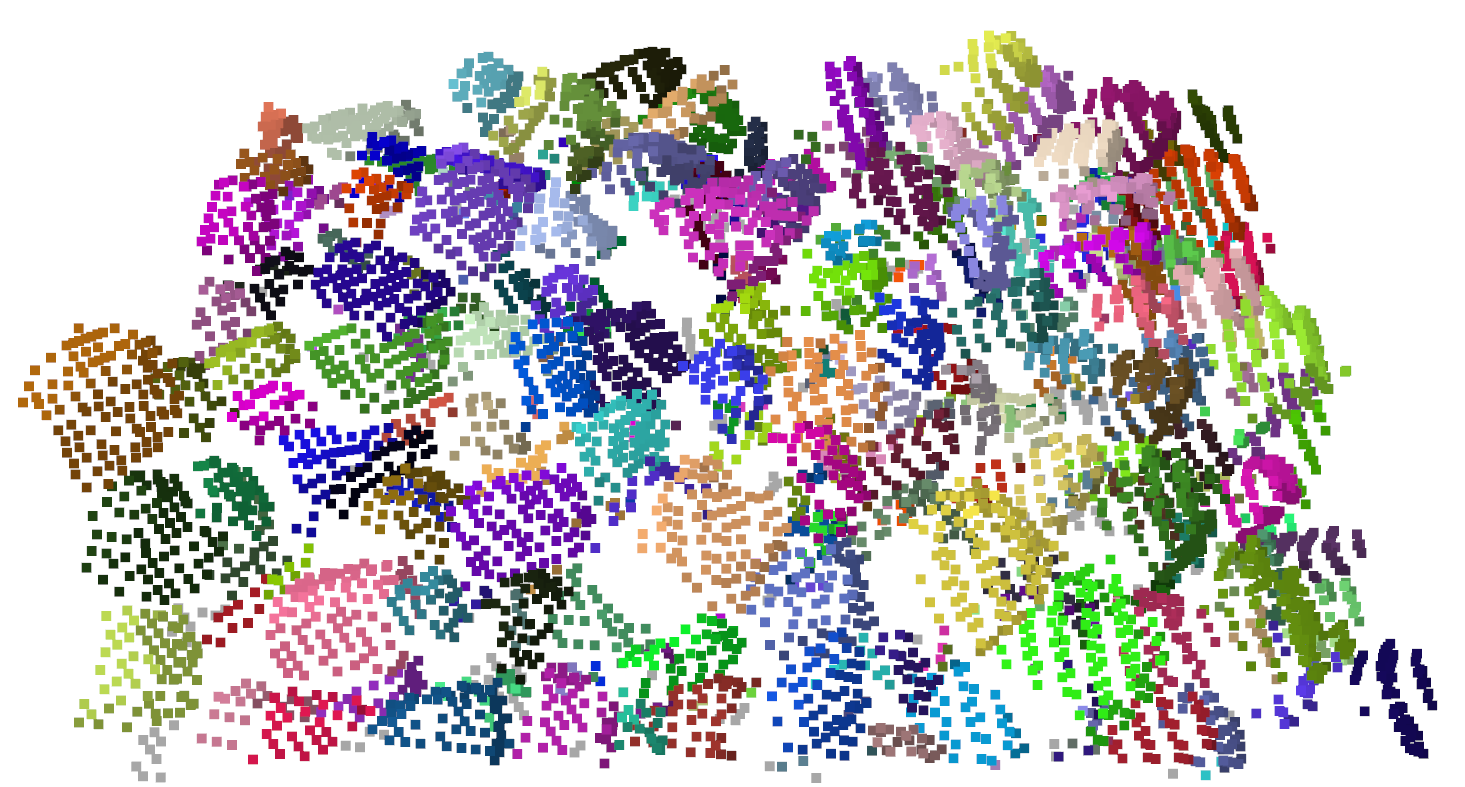}}}\put(-67,50){\scalebox{1.0}{Simulation}} \vline&
{\raisebox{0.\height}{\includegraphics[trim=0cm 0cm 0cm 0cm,clip,width=.18\linewidth]{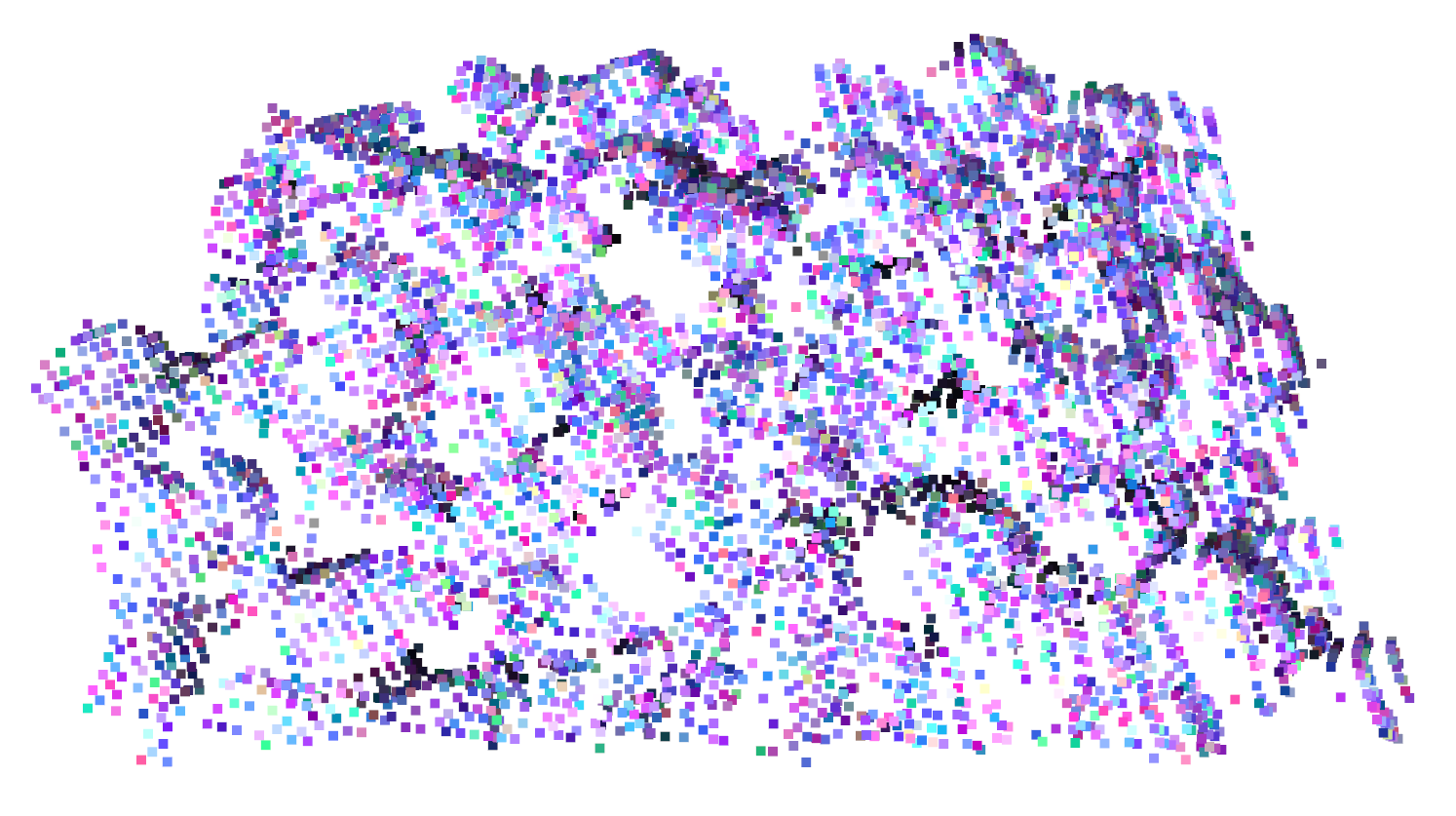}}}\put(-17,50){\scalebox{1.0}{Normal}} &
{\raisebox{0.\height}{\includegraphics[trim=0cm 0cm 0cm 0cm,clip,width=.18\linewidth]{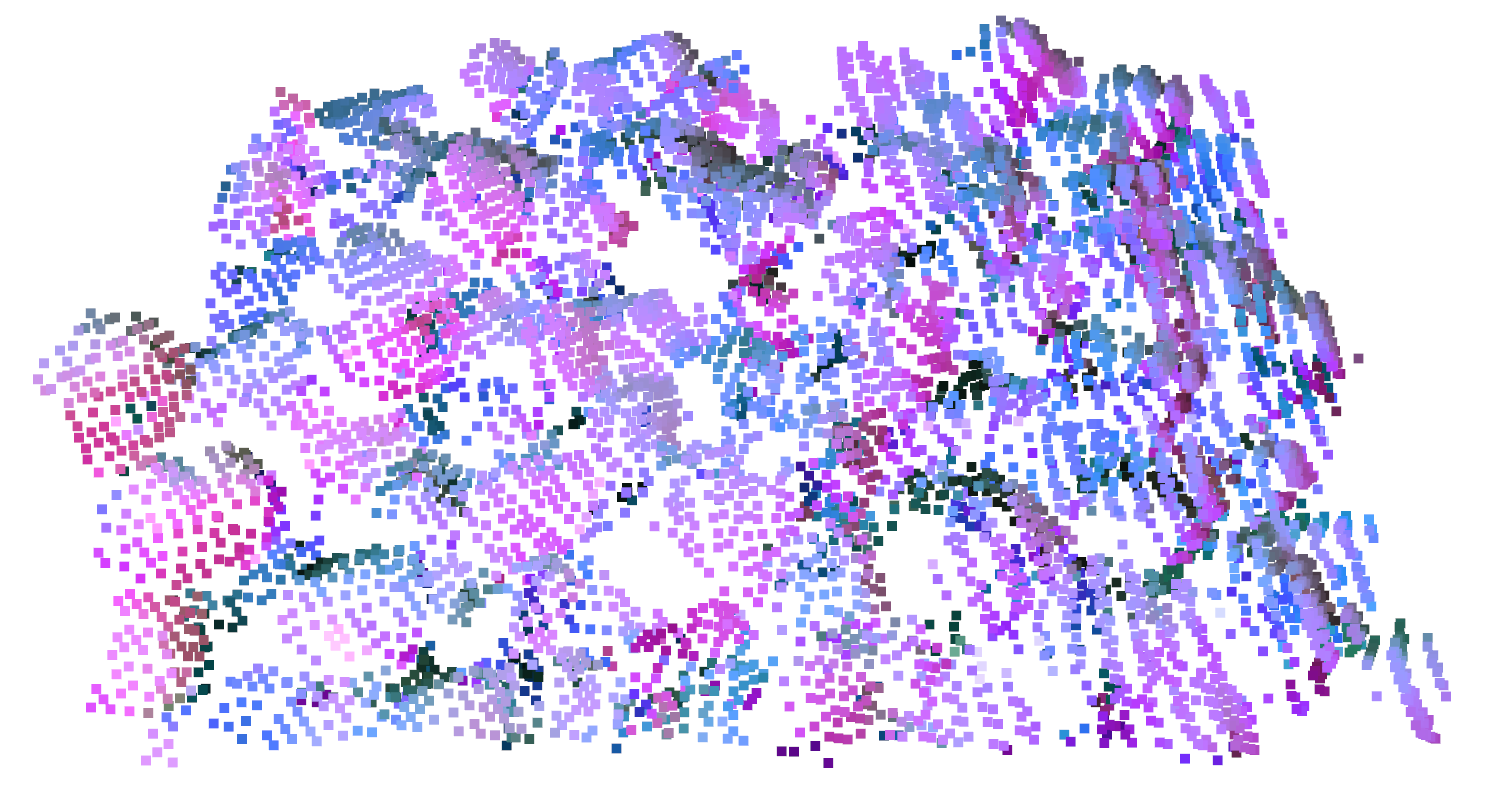}}}\vline&
{\raisebox{0.\height}{\includegraphics[trim=0cm 0cm 0cm 0cm,clip,width=.18\linewidth]{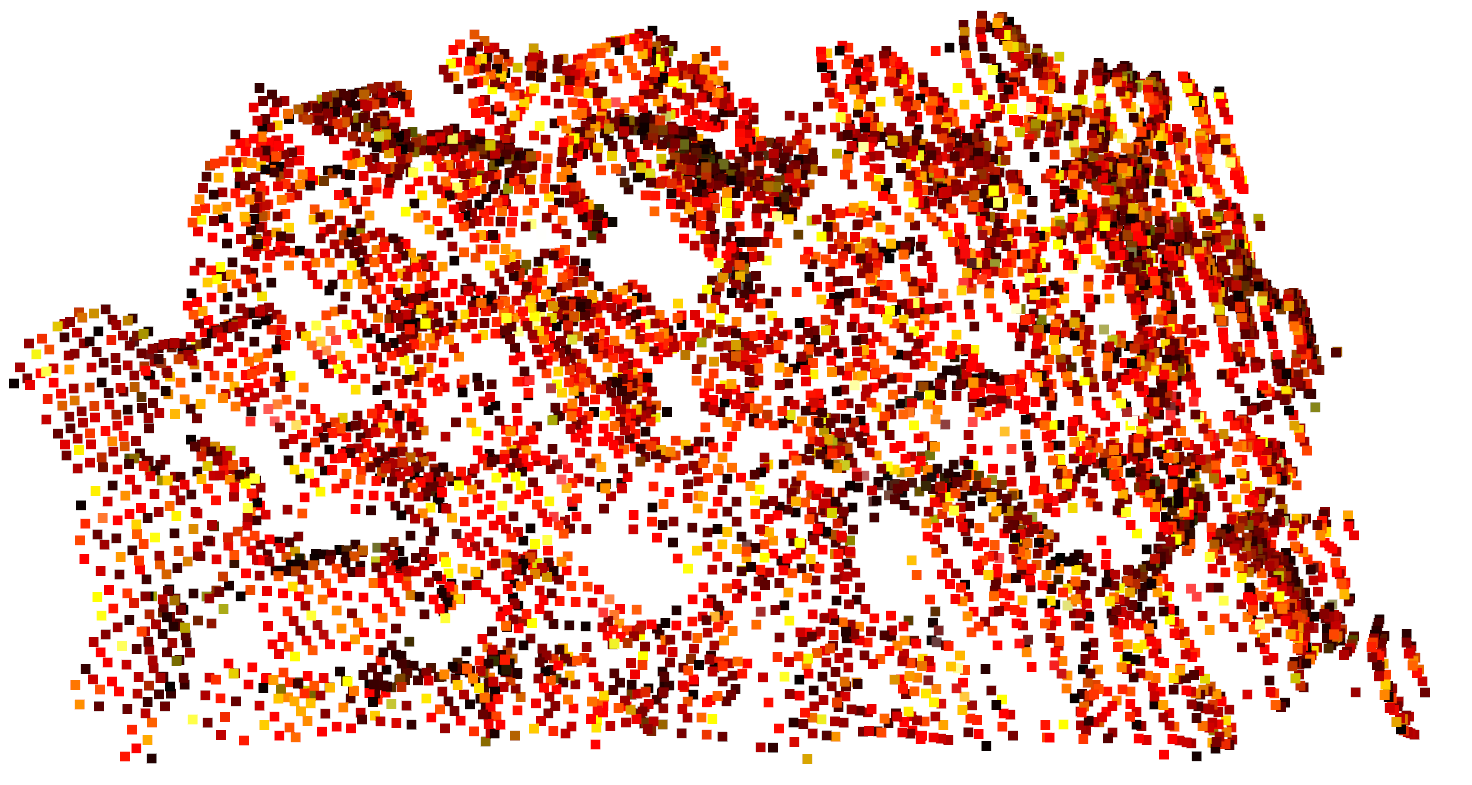}}}\put(-21,50){\scalebox{1.0}{Curvature}} &
{\raisebox{0.\height}{\includegraphics[trim=0cm 0cm 0cm 0cm,clip,width=.18\linewidth]{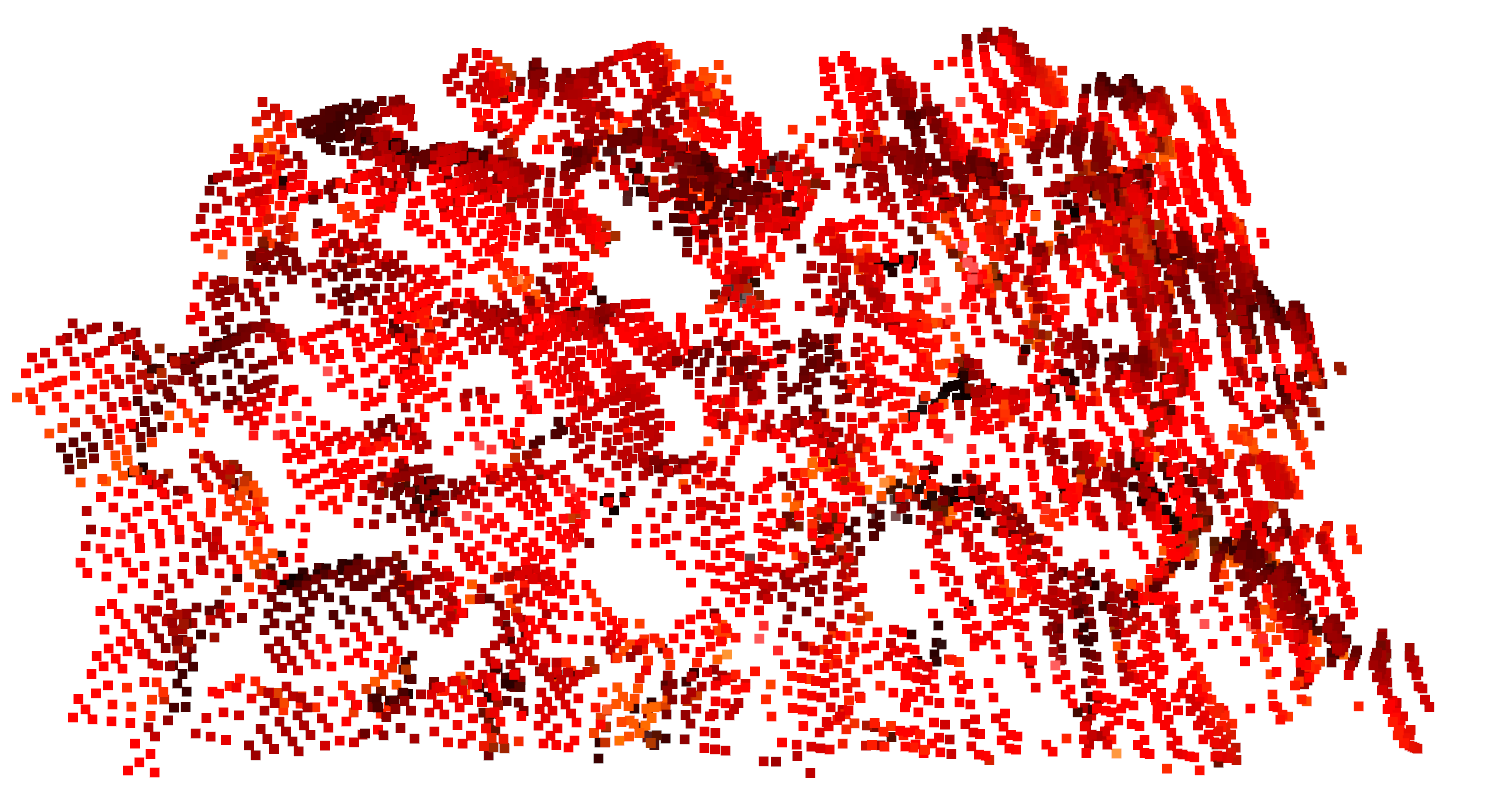}}}\put(-50,50) \vline &
\raisebox{-0.5\height}[0pt][0pt]{\includegraphics[trim=0cm 0cm 0cm 0cm,clip,width=.06\linewidth]{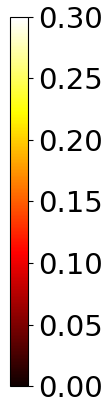}} \\
{\raisebox{0.\height}{\includegraphics[trim=0cm 0cm 0cm 0cm,clip,width=.18\linewidth]{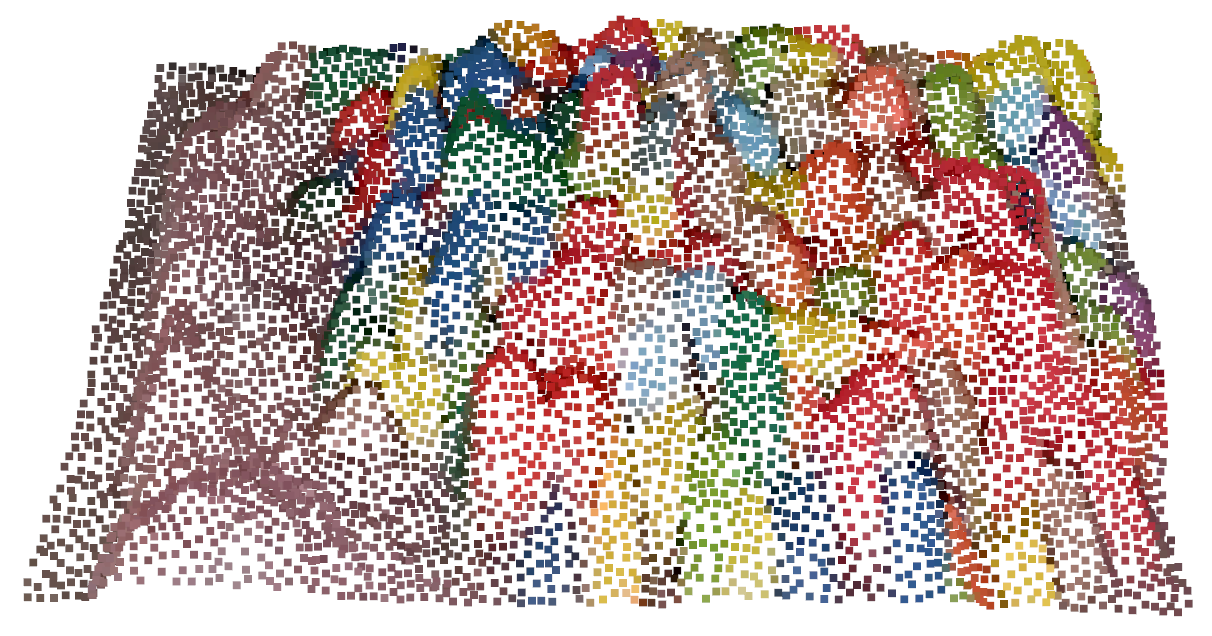}}}\put(-68,47){\scalebox{1.0}{Real World}} \vline&
{\raisebox{0.\height}{\includegraphics[trim=0cm 0cm 0cm 0cm,clip,width=.18\linewidth]{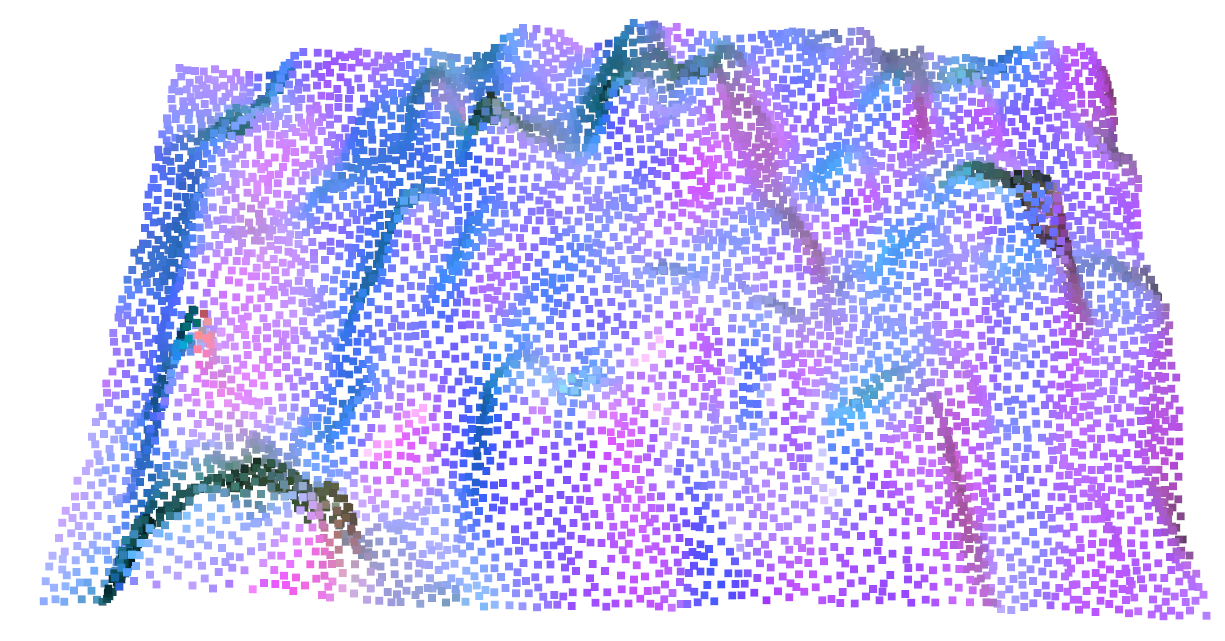}}}\put(-68,-10){\scalebox{0.8}{Ground Truth}} &
{\raisebox{0.\height}{\includegraphics[trim=0cm 0cm 0cm 0cm,clip,width=.18\linewidth]{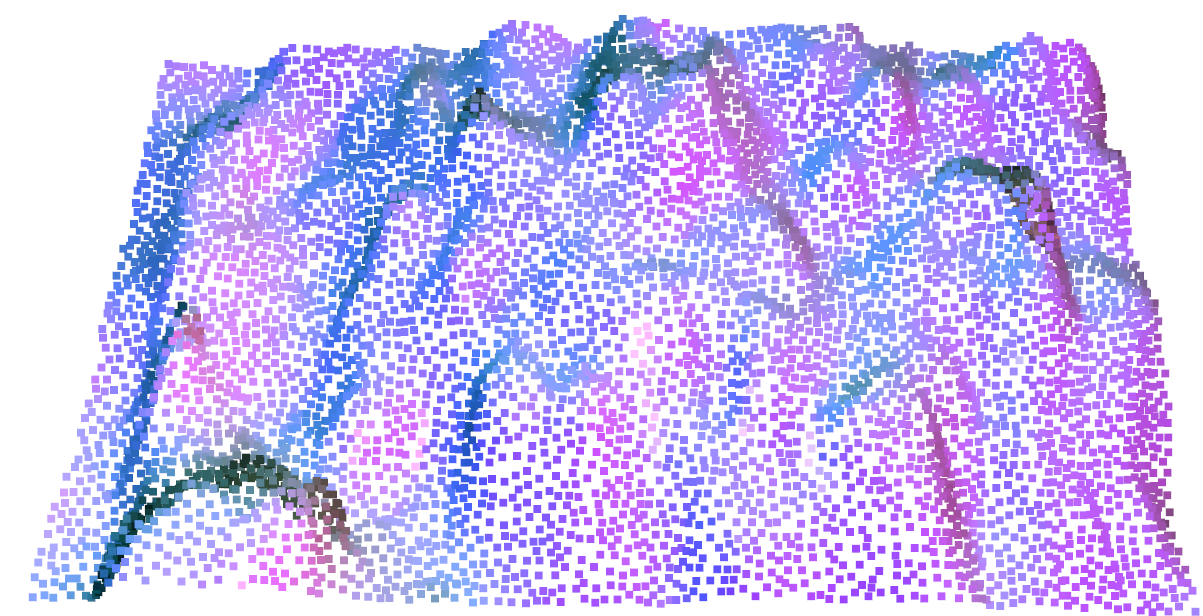}}}\put(-61,-10){\scalebox{0.8}{Prediction}} \vline&
{\raisebox{0.\height}{\includegraphics[trim=0cm 0cm 0cm 0cm,clip,width=.18\linewidth]{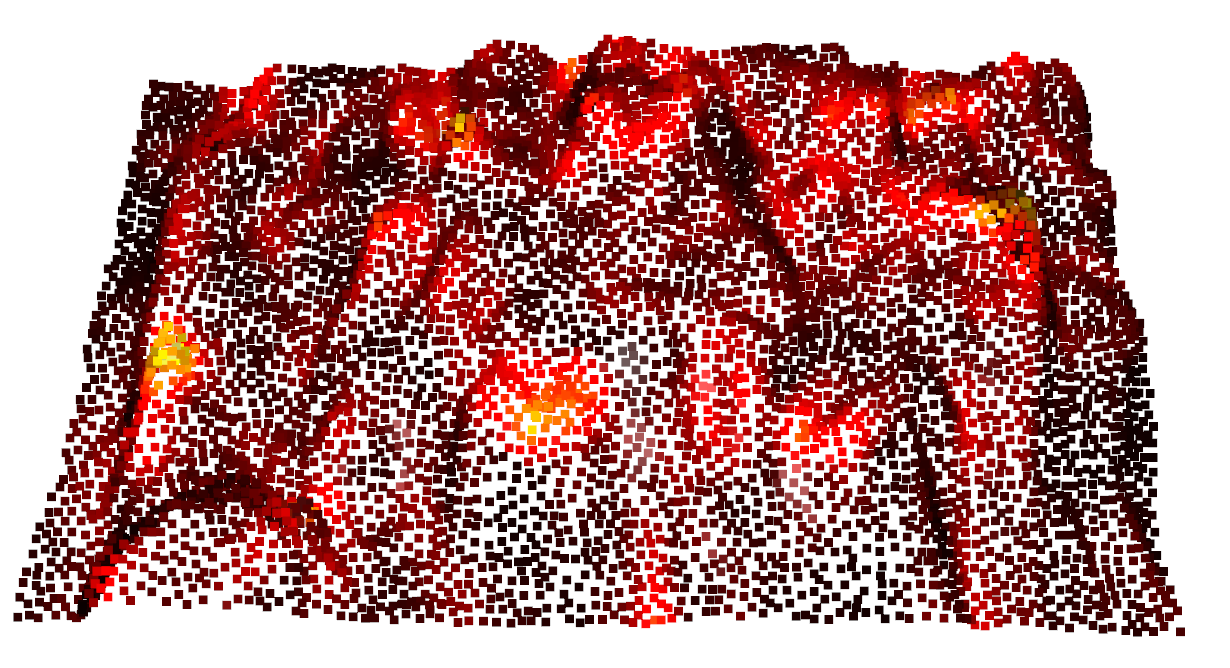}}}\put(-68,-10){\scalebox{0.8}{Ground Truth}} &
{\raisebox{0.\height}{\includegraphics[trim=0cm 0cm 0cm 0cm,clip,width=.18\linewidth]{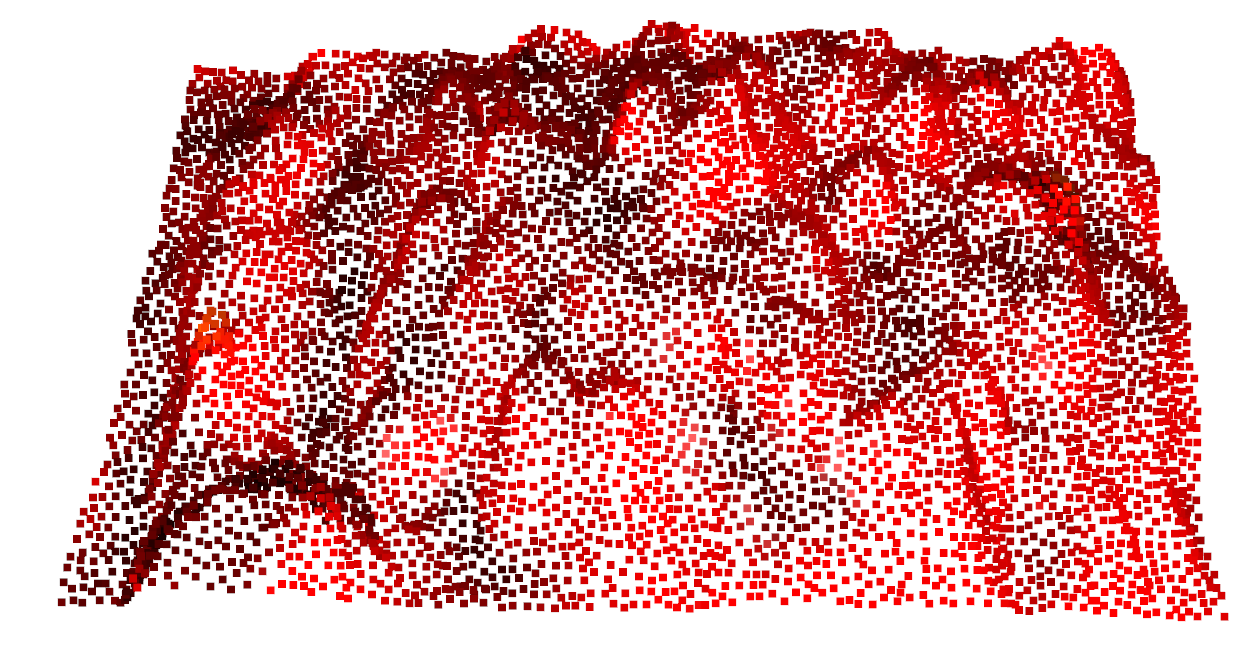}}}\put(-61,-10){\scalebox{0.8}{Prediction}}  
\end{tabular}
\caption{The ground truth and predictions from the learned representation for both simulated and physical scenes. In the pictures of normals, the normal legend is: \textcolor{red}{red $\rightarrow$ X}; \textcolor{green}{green $\rightarrow$ Y}; \textcolor{blue}{blue $\rightarrow$ Z}. The point clouds for curvature are colored with curvature values at the points, with the scale shown on the far right. \label{fig:norm_curv}}
\end{figure*}

We collect approximately 35,000 unique scenes of rigid objects excavation in simulation, and compute the ground truth normal and principal curvature of the point clouds with the Point Cloud Library~\footnote{https://pointclouds.org/}. 
We downsample each point cloud to 7,000 points before training. We further augment the data by introducing random rigid translations to the entire point cloud. We split the data into 90\% of training data and 10\% of testing data, and train the model. We adopt a batch size of 16, the Adam optimizer, learning rate of 0.01 with a $1e^{-4}$ decay rate, and a total of 10 epochs. The testing results are summarized in Table.~\ref{table:rep_res}. 
The learned model predicts the normal with a mean absolute error (MAE) of $22.0^{\circ}$ in simulation and $15.4^{\circ}$ in the real world. The curvature prediction MAE is around $0.03$ for both simulated and real data. The ground truth principle curvature ranges from 0 to $0.33$.
The object counting MAE in simulation is $16.1$ and the maximum objects number is $300$. 
Qualitatively, the learned model predicts the normal and curvature well for point clouds in simulation and the real world, as shown in Fig.~\ref{fig:norm_curv}. 

We predict the normal, curvature, and objects number directly on the physical point clouds without any tuning. We perform evaluations on a total of 25 unique scenes and the results are also shown in Table~\ref{table:rep_res}. Surprisingly, even without any tuning on the real scene, the normal MAE is smaller than that in simulation. Upon a closer look at the qualitative results in Fig.~\ref{fig:norm_curv}, we find that the real world point cloud is quite smooth, which is an artifact of the Azure RGB-D sensor. As a result, the model is able to capture the normal and curvature of real-world point clouds even better than those in simulation. The average predicted number of objects in the 25 scenes is 131. We did not obtain the ground truth object counts to calculate the MAE because the sizes of the objects in the tray are smaller than those in simulation on average, which makes quantitative object counting results less meaningful. However, we notice in these 25 scenes, the predicted number of objects would decrease when the objects are excavated from the scene in the real world.

\begin{table}[h]
\centering
    \caption{\label{table:rep_res} Representation Validation Results}
    \begin{tabular}{c@{\hspace{0.35\tabcolsep}}|@{\hspace{0.35\tabcolsep}}c@{\hspace{0.35\tabcolsep}}|@{\hspace{0.35\tabcolsep}}c@{\hspace{0.35\tabcolsep}}|@{\hspace{0.35\tabcolsep}}c@{\hspace{0.35\tabcolsep}}c}
    \toprule
      \textbf{Environment} & \textbf{Normal Loss}&\textbf{Curv. MAE} & \textbf{Obj. MAE}\\
      \hline
     Simulation & -0.927 / 22.0$^{\circ}$ & 0.0314 &  16.1 \\
      \hline
     Real World & -0.964 / 15.4$^{\circ}$ & 0.0323 & - \\
    \bottomrule
    \end{tabular}
\end{table}
\vspace{-5pt}

\vspace{-5pt}
\vspace{-5pt}





\subsection{RL Training}
We perform RL excavation training in simulation using the PPO algorithm from Stable Baseline3~\footnote{https://github.com/DLR-RM/stable-baselines3}. In our method (denoted as RL-rep-exv), we use the learned representation to extract the compact features, and feed into the policy network with 2 fully connected layers of 256 and 64 ReLU neurons. The compact geometric representation allows us to train a simple fully-connected network with only 16k parameters. The policy is updated every 768 samples with 6 simulations running in parallel. 

We also compare with a RL approach without using the representation, denoted as RL-exv. 
For the RL-exv policy network, we stack two fully-connected layers on top of the feature extraction layers. Its fully-connected layers have the same architecture with the policy network of RL-rep-exv and the feature extraction component is the same as the representation model. RL-exv directly takes the point cloud after FPS as the input. Note both the feature extraction layers and fully-connected layers are updated during training. The entire policy network of RL-exv has 205k parameters that need to be updated during training. 



\begin{figure}[h]
    \vspace{-15px}
    \centering
    \includegraphics[width=0.4\textwidth]{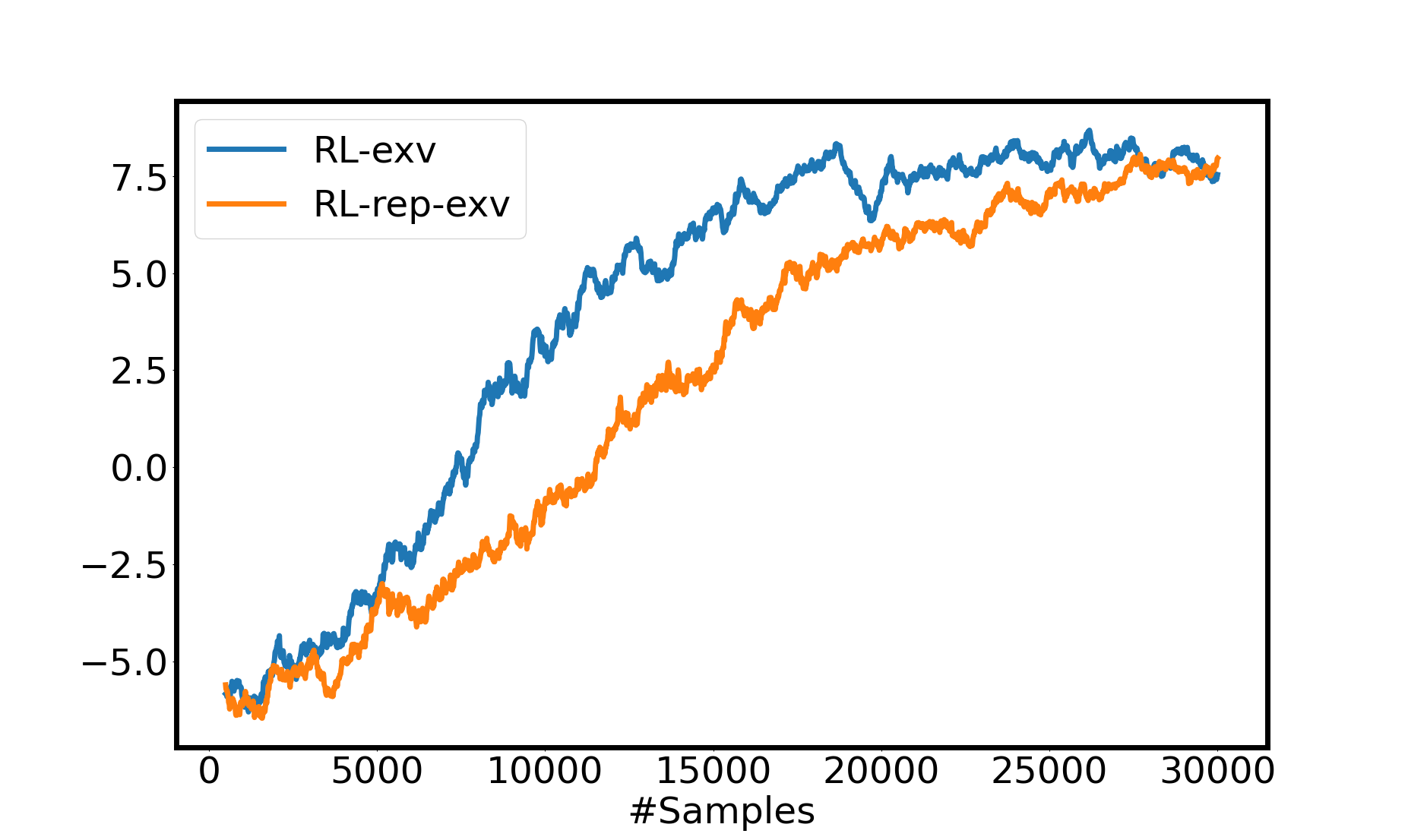}
    \caption{The training curves of RL-exv and RL-rep-exv. The y and x axes represent the RL reward and the number of training samples.}
    \label{fig:RL_training}
\end{figure}

To guide the RL agent to learn successful trajectory planning, if the planning fails, a -1 reward is given instead of 0. A discount factor of $0.99$ is used for the excavation learning. The action parameters are scaled between $-1$ and $1$ and the rewards are normalized by dividing the standard deviation of historical rewards. For RL in simulation, at the start of an episode, we randomly select a random number (between 200 and 300) of training objects and spawn them into the digging tray with random object poses to create an excavation scene. The RL agent excavates 10 times for each episode. Both RL-rep-exv and RL-exv are trained with 30,000 samples (i.e., 3000 episodes) in simulation. The training reward curves of both methods are shown in Fig.~\ref{fig:RL_training}. Although RL-rep-exv converges slower than RL-exv during training, it achieves similar asymptotic performance with RL-exv using significantly fewer input features.  

We perform the representation learning, RL training, and experiments on an desktop computer with an Xeon Silver 4114 processors, 32GB RAM, and two Nvidia GeForce GTX 1080 graphics cards.
It takes 126 hours 39 minutes and 163 hours 28 minutes to train RL-rep-exv and RL-exv respectively. Due to our compact geometric visual representation, RL-rep-exv trains much faster than RL-exv.


\subsection{Excavation Experiments in Simulation}


For excavation experiments in simulation, we select a random number (between 50 and 300) of objects to create an excavation scene for each RL episode, and add to the digging tray same as training. The RL agent also plans and executes 10 excavations sequentially for each episode. We experiment with 100 episodes for both RL-rep-exv and RL-exv. We set the penetration depth $d = 0.1$\,m, the maximum dragging length $d_{max} = 0.3$\,m, and the fixed closing angle $\beta = 135^\circ$. We always lift the bucket to $0.4$\,m from the base during lifting.

We benchmark the excavation experiments of both methods using the average excavation volume (Avg. V) and the trajectory planning success rate (Plan Succ. R.) of all 1000 experimented excavations. We also measure the average excavation volume of excavations (Avg. V w/ Plan) with valid trajectories. The trajectory planning failures are caused by IK failure or collisions, as discussed in Section~\ref{sec:problem_define}. 

We compare the two RL methods with two scripted baseline approaches, namely ``heuristic" and ``random". The heuristic planner always selects the highest attacking point. The random planner randomly select an attacking point in the tray for each excavation. The excavation angles are uniformly sampled from the excavation angle range mentioned in Section\mbox{\ref{sec:sim_setup}} of the appendix for both baseline methods. Excavations are sequentially planned and executed for each episode of the baseline methods until there are $10$ excavations with valid trajectories.

\begin{figure}[h]
    \centering
    \subfloat[\centering RL-rep-exv attacking points.]{{\raisebox{0.02\height}{\includegraphics[trim=0cm 0cm 0cm 0.cm,clip,width=.48\linewidth]{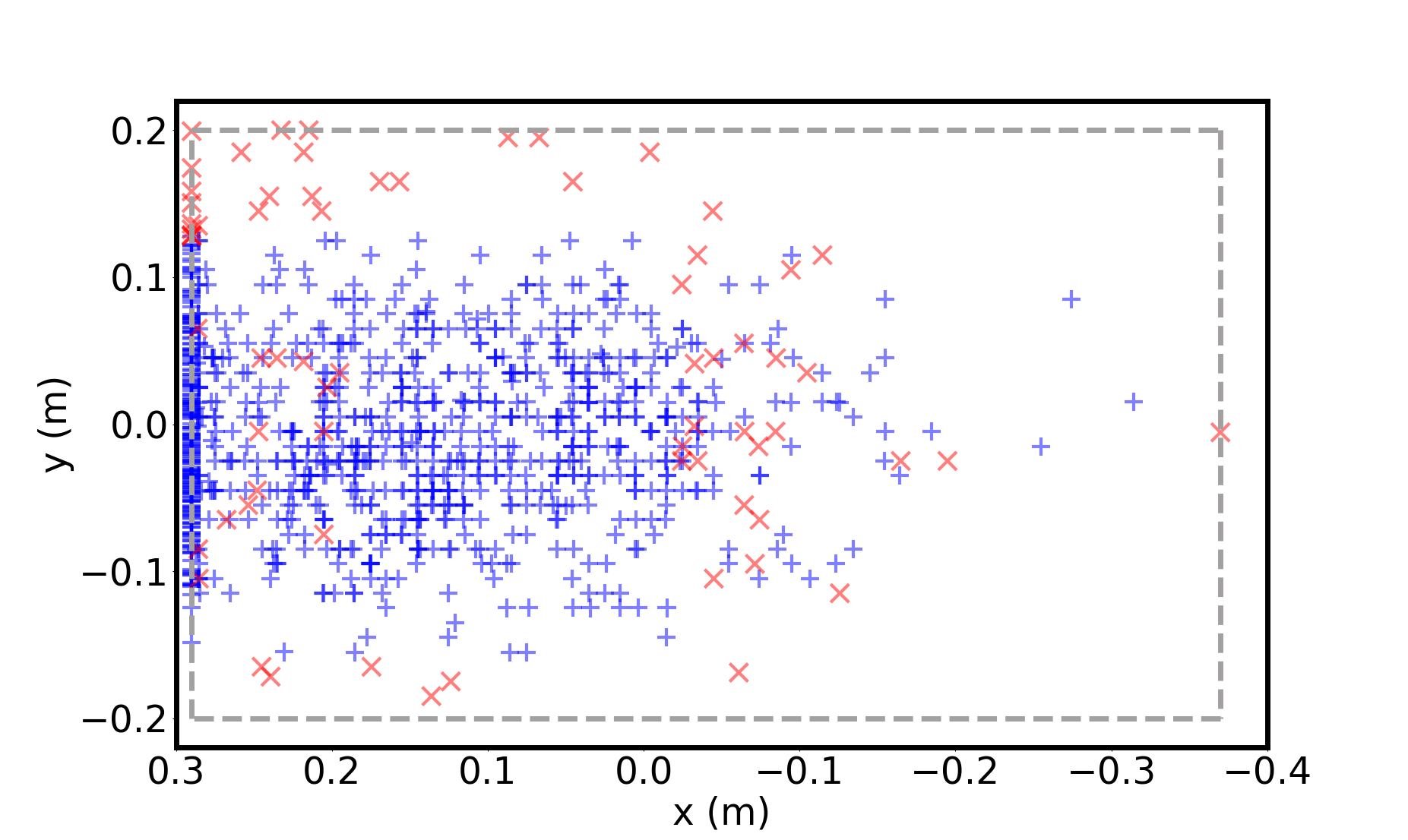}}}\label{fig:RL_rep_sim_poa}}
    \subfloat[\centering RL-exv attacking points.]{{\includegraphics[trim=0cm 0cm 0cm 0cm,clip,width=.4825\linewidth]{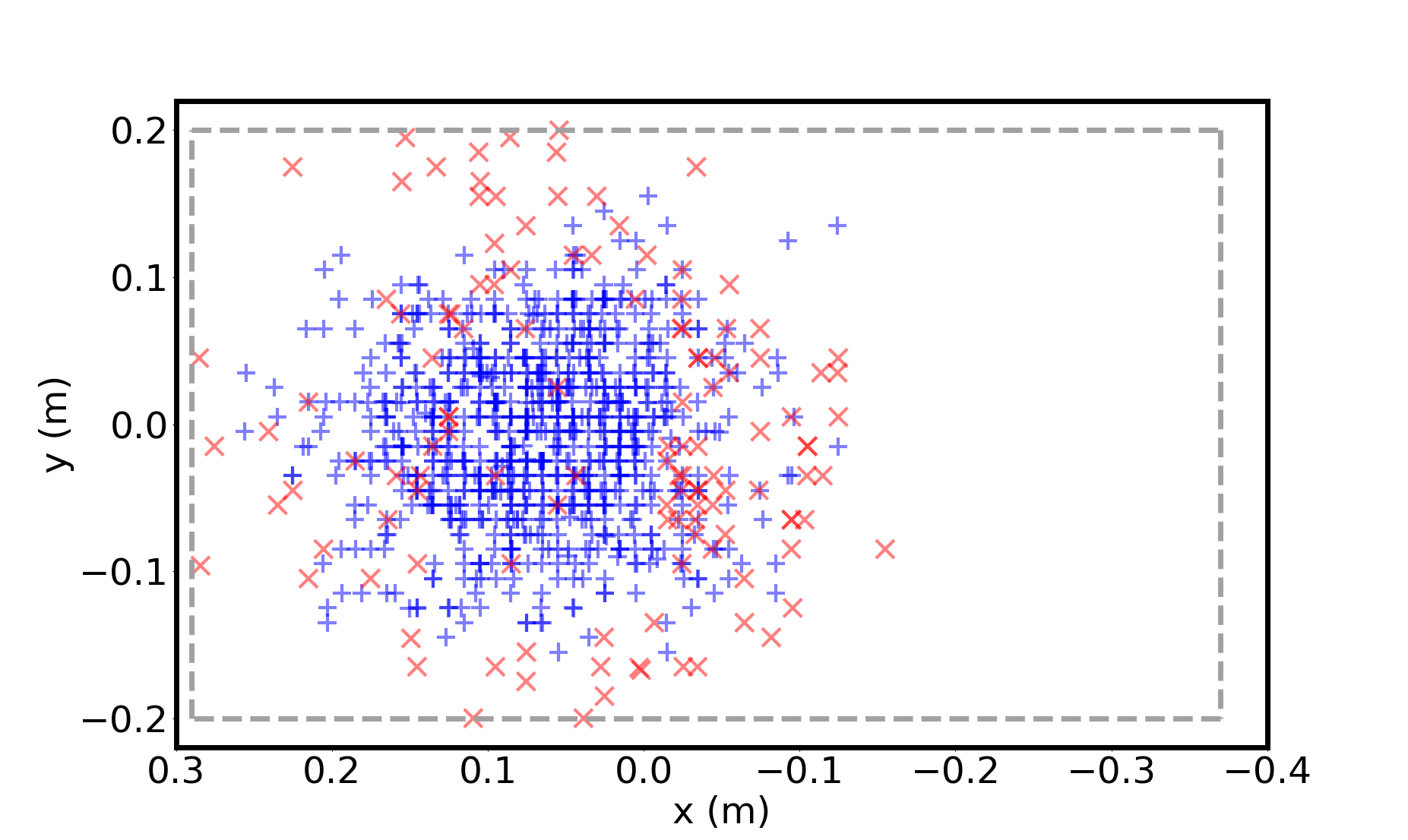} }\label{fig:RL_sim_poa}}
    \caption{The attacking 2D points of both methods for simulated experiments are plotted in this figure. The attacking points with valid plans are colored blue and the ones without plans are colored red. The dashed grey box shows the excavation range in the tray.}
    \label{fig:sim_PoA}
    \vspace{-2mm}
\end{figure}

\begin{figure*}[]
\centering
\setlength{\tabcolsep}{0px}
\begin{tabular}{lllll}
{\raisebox{0.0\height}{\includegraphics[trim=6.5cm 5cm 4cm 4cm,clip,width=.19\linewidth]{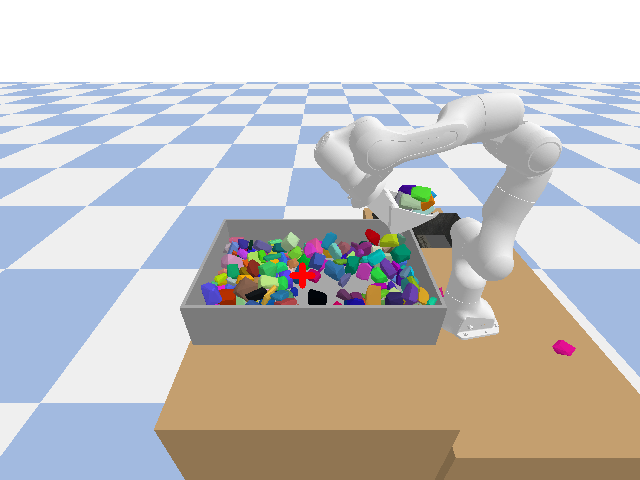}}} &
{\raisebox{0.\height}{\includegraphics[trim=6.5cm 5cm 4cm 4cm,clip,width=.19\linewidth]{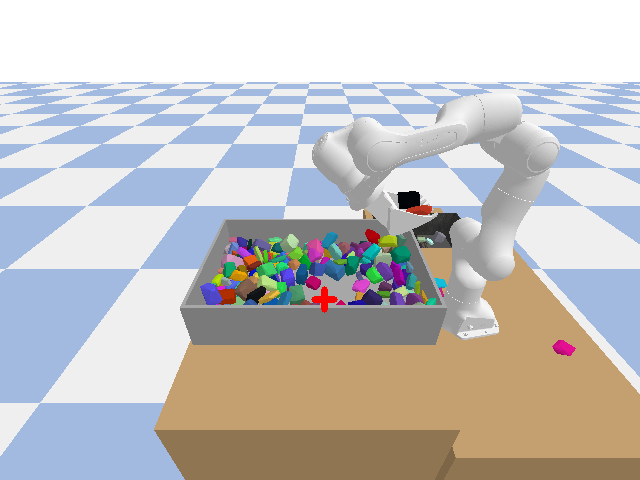}}} &
{\raisebox{0.\height}{\includegraphics[trim=6.5cm 5cm 4cm 4cm,clip,width=.19\linewidth]{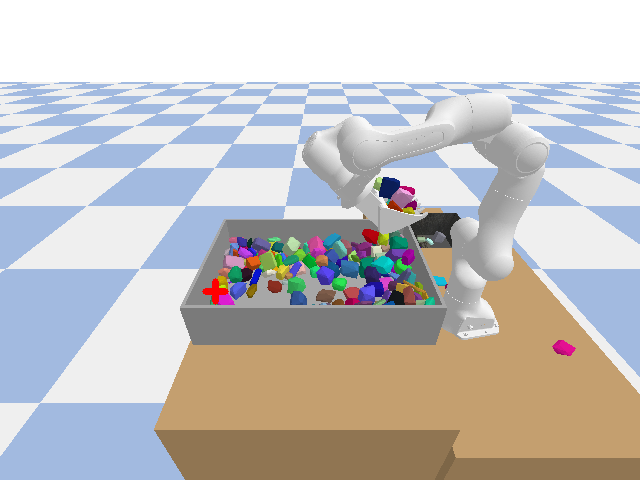}}} &
{\raisebox{0.\height}{\includegraphics[trim=6.5cm 5cm 4cm 4cm,clip,width=.19\linewidth]{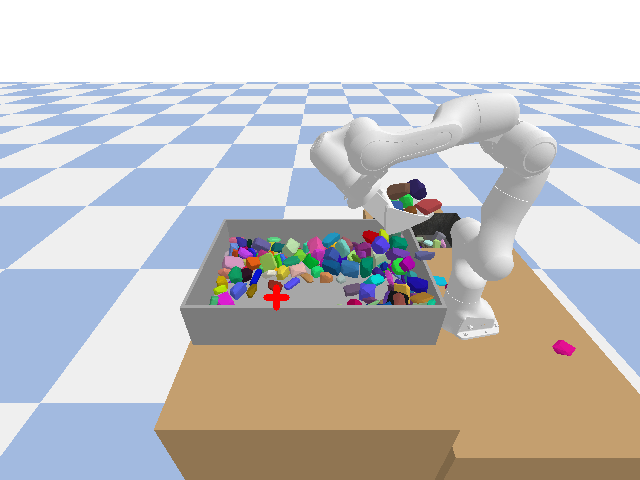}}} &
{\raisebox{0.\height}{\includegraphics[trim=6.5cm 5cm 4cm 4cm,clip,width=.19\linewidth]{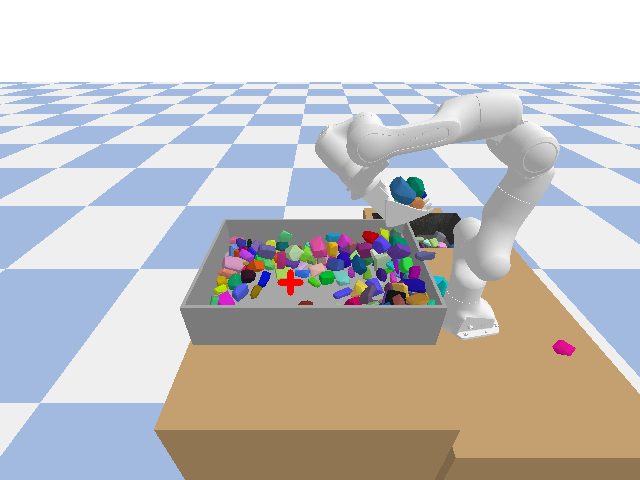}}} \\
{\raisebox{0.\height}{\includegraphics[trim=6.5cm 5cm 4cm 4cm,clip,width=.19\linewidth]{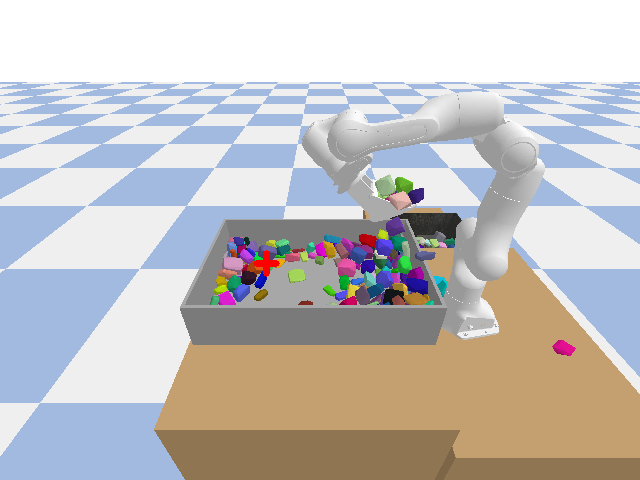}}} &
{\raisebox{0.\height}{\includegraphics[trim=6.5cm 5cm 4cm 4cm,clip,width=.19\linewidth]{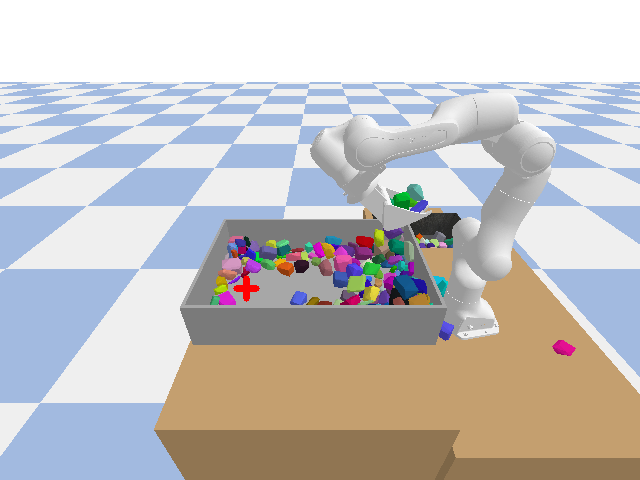}}} &
{\raisebox{0.\height}{\includegraphics[trim=6.5cm 5cm 4cm 4cm,clip,width=.19\linewidth]{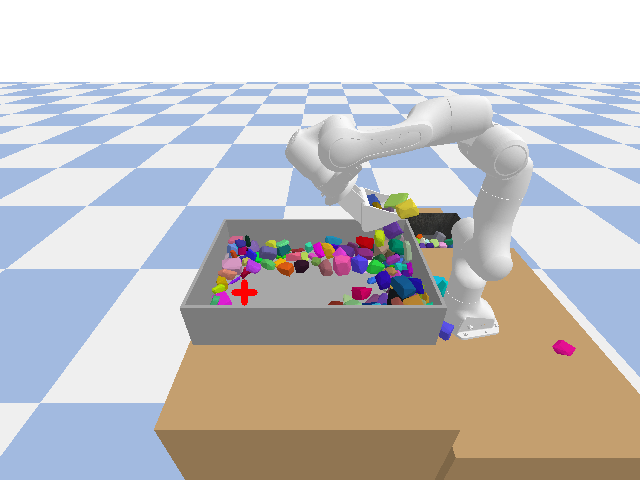}}} &
{\raisebox{0.\height}{\includegraphics[trim=6.5cm 5cm 4cm 4cm,clip,width=.19\linewidth]{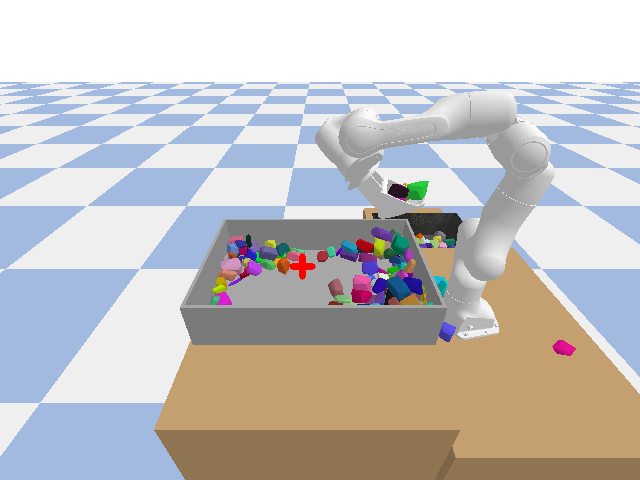}}} &
{\raisebox{0.\height}{\includegraphics[trim=6.5cm 5cm 4cm 4cm,clip,width=.19\linewidth]{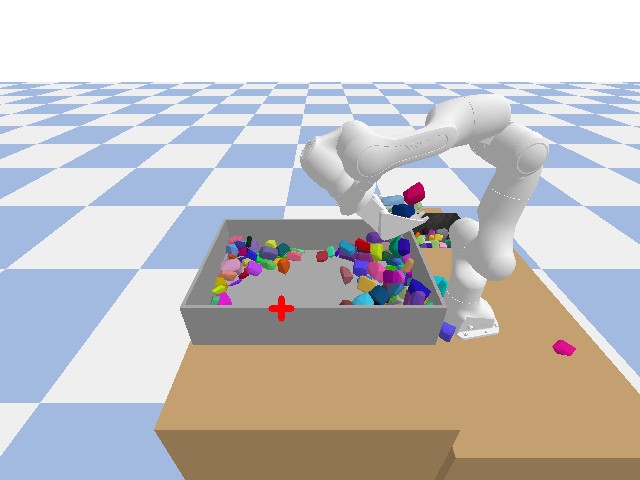}}} 
\end{tabular}
\vspace{5px}
\caption{We qualitatively show the excavation results of one complete episode for the simulated experiments of RL-rep-exv. The red cross in each image represents its planned 2D attacking point. View order: left to right and top to bottom.}
\label{fig:sim_sequence}
\end{figure*}

As shown in Table~\ref{table:sim_rl_exp}, RL-rep-exv outperforms RL-exv in terms of all three metrics for simulated experiments, demonstrating that the learned representation transfers better to the physical scene. We visualize the experimented 2D attacking points (without attack angle) for both methods in Figure~\ref{fig:sim_PoA}. Compared with RL-exv, RL-rep-exv generates attacking points further away from the robot base, which allows it to drag longer and potentially dig more objects during excavation. The mean and standard deviation of the attack angles are 28.9$^\circ$ and 23.6$^\circ$ for RL-rep-exv, and 43.1$^\circ$ and 27.4$^\circ$ for RL-exv, respectively.
In Figure~\ref{fig:sim_sequence}, we qualitatively show 10 sequential excavations together with their corresponding attacking points for one complete experimented episode of RL-rep-exv. As can be seen this example, the trained policy of RL-rep-exv plans attacking points in order to maximize the volume of excavated objects.

RL-rep-exv and RL-exv significantly outperform the two heuristic approaches in terms of the average volumes and the planning success rate. The average volume with plan of RL-rep-exv is larger than the that of the random planner. However, the heuristic planner achieves the highest average volume with plan and outperforms both RL methods. The comparison with baseline methods means RL learns to significantly reduce the planning failures caused by IK failures and collision. However, both RL methods are not able to improve the average volume with plans effectively. We are interested to further study and improve the RL performance for excavations with valid plans.

\begin{table}[h]
\centering
    \caption{\label{table:sim_res} Simulated experimental results (the bucket filling rate shown in parentheses)}
    \begin{tabular}{c@{\hspace{0.35\tabcolsep}}|@{\hspace{0.35\tabcolsep}}c@{\hspace{0.35\tabcolsep}}|@{\hspace{0.35\tabcolsep}}c@{\hspace{0.35\tabcolsep}}|@{\hspace{0.35\tabcolsep}}c@{\hspace{0.35\tabcolsep}}c}
    \toprule
      \textbf{Method} & \textbf{Avg. V}&\textbf{Plan Succ. R.} & \textbf{Avg. V w/ Plan}\\
      \hline
     RL-rep-exv & 194.1 (43.1\%) & 93.1\% &  208.4 (46.3\%) \\
      \hline
     RL-exv & 176.6 (39.2\%)& 87.4\% & 202.1 (44.9\%)\\
     \hline
     Heuristic & 95.1 (21.1\%) & 39.5\% & 240.8 (53.5\%) \\
        \hline
     Random & 68.5 (15.2\%) &33.7\% & 203.4 (45.2\%) \\
    \bottomrule
    \end{tabular}

     
\label{table:sim_rl_exp}
\end{table}
\vspace{-2mm}

\subsection{Excavation Experiments in Real World}

We directly deploy the RL policies trained in simulation for excavation experiments in the real world. We create a excavation scene by shaking the $282$ rigid objects in a box, and then pouring them into the excavation area of the tray. The robot dumps the excavated objects into a dumping tray after each excavation. 
5 excavations are planned and executed sequentially by the RL agent for each episode. It can be hard to dig objects after 5 excavations in real world, because the objects clutter becomes more compact and there are fewer objects left after each excavation. So we perform 5 excavations for each episode in the real world, instead of 10 as simulation. We experiment with 5 episodes (i.e., 25 excavations) for both RL-rep-exv and RL-exv. Different from simulation, we set the penetration depth $d = 0.07$\,m, the maximum dragging length $d_{max} = 0.1$\,m to account for the limited power of the physical robot arm. 

We evaluate the physical excavation experiments of both methods using the average excavation volume (Avg. V) and the trajectory planning success rate (Plan R.) of all 25 experimented excavations. Moreover, we measure the excavation trajectory execution success rate (Exe. R.) and the average volume of the successful excavations (Avg. V w/ Succ. Exe.).

\begin{figure}[h!]
    \centering
    \includegraphics[trim=0cm 0cm 0cm 1cm, clip,width=0.4\textwidth]{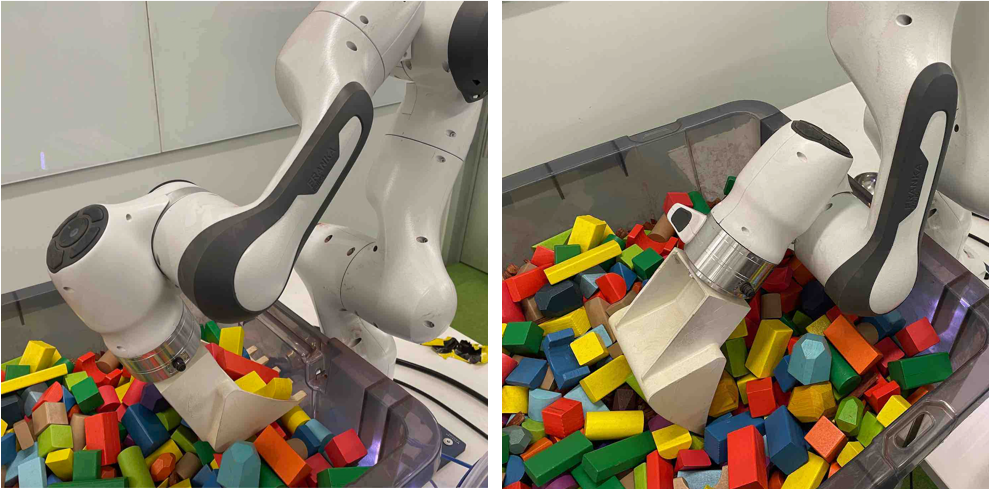}
    \caption{Two typical failure cases of excavation planning due to collision with the tray are shown in the figure. The collisions in the left and right images happen during the closing and dragging phases respectively.}
    \label{fig:plan_fail_example}
\end{figure}

As seen in Table~\ref{table:real_exv_res}, RL-rep-exv outperforms RL-exv in terms of average excavation volume and average volume of successfully executed excavations for physical experiments. Moreover, all $25$ attacking poses generated by RL-rep-exv lead to valid joint trajectories, while 8 out of 25 excavations fail to generate joint trajectories for RL-exv.  Two typical environment collision failure cases during planning are shown in Figure~\ref{fig:plan_fail_example}. 
The two observations demonstrate the benefits of the excavation RL using the geometric representation. 
RL-rep-exv achieves an average excavation volume of $191$ cm$^{3}$. The average excavation volume of successfully executed excavations is $341.1$ cm$^{3}$ for RL-rep-exv. 
There are fewer objects to dig after each excavations, which makes it harder for later excavations to achieve high excavation volumes. Since one episode contains different number of excavations for simulated and physical experiments (5 and 10 respectively), we can not directly compare the excavation volumes between simulation and the real world. 

\begin{figure}[h]
    \centering
    \subfloat[\centering RL-rep-exv PoA]{{\raisebox{0.0\height}{\includegraphics[trim=0cm 0cm 0cm 0.cm,clip,width=.48\linewidth]{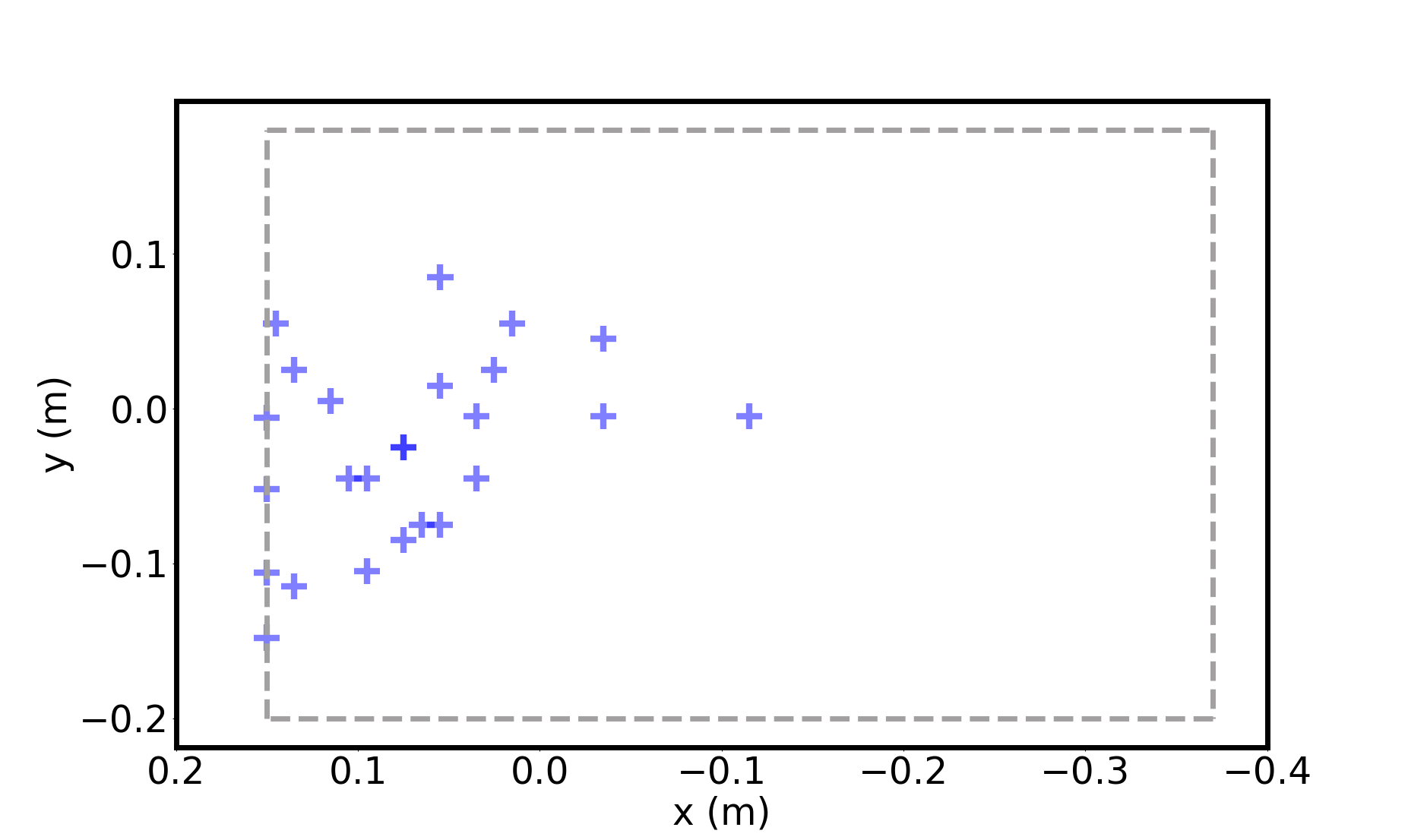}}}\label{fig:RL_rep_real_poa}}
    \subfloat[\centering RL-exv PoA]{{\includegraphics[trim=0cm 0cm 0cm 0cm,clip,width=.4825\linewidth]{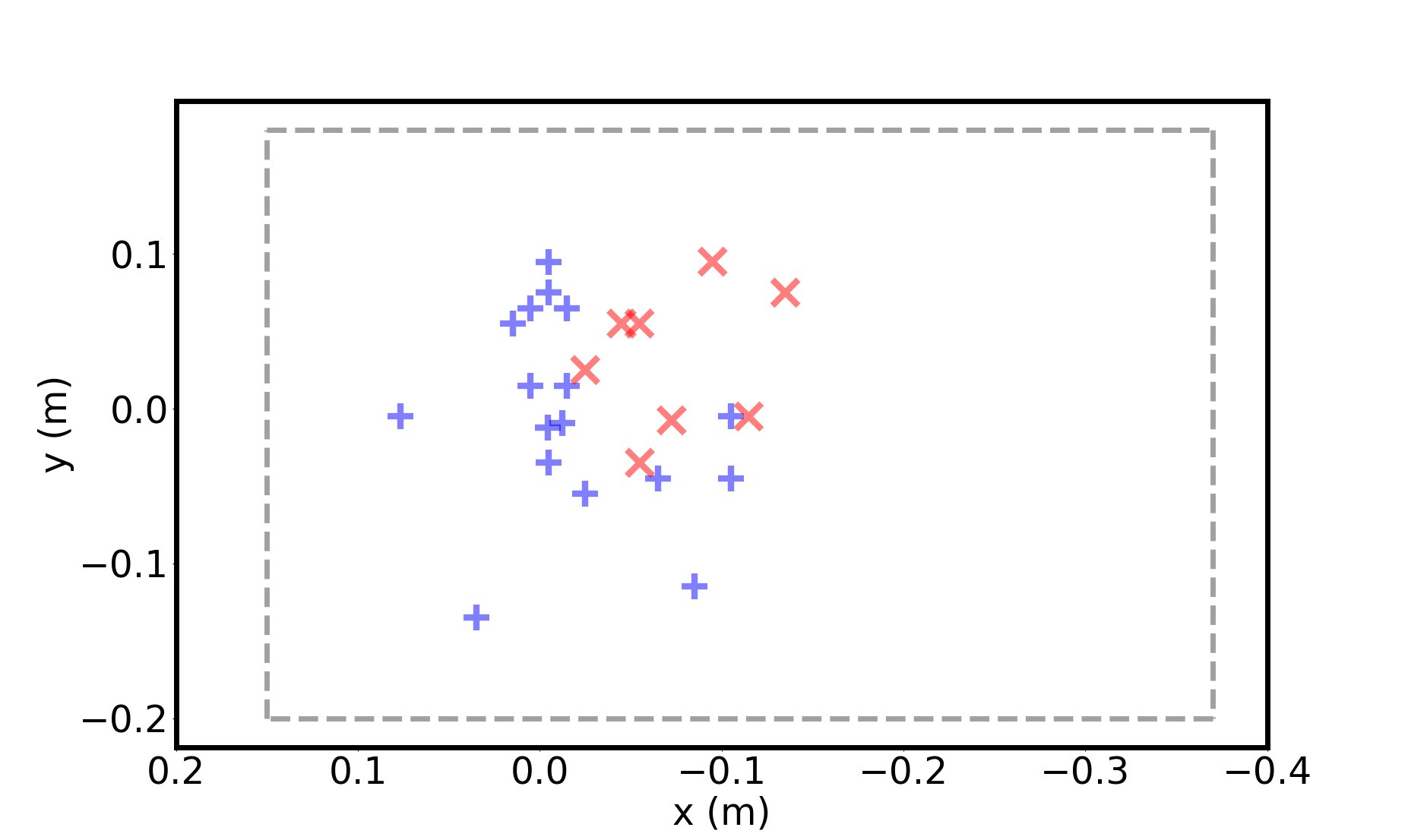} }\label{fig:RL_real_poa}}
    \caption{The attacking 2D points of both methods for physical experiments are visualized in this figure. The attacking points with valid plans are colored blue and the ones without plans are colored red. The dashed grey box shows the excavation range in the tray.}
    \label{fig:real_PoA}
\end{figure}

\begin{figure}[h!]
    \centering
    \includegraphics[width=0.45\textwidth]{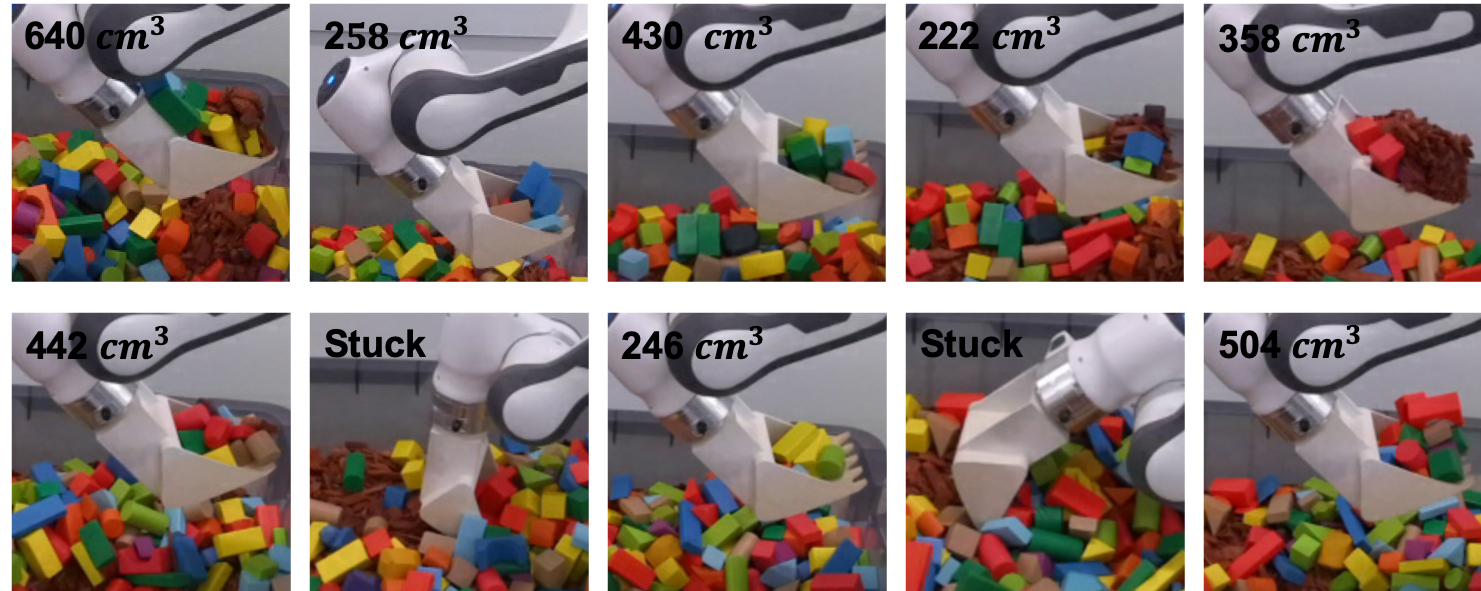}
    \caption{The two rows of images show two excavation episodes for RL-rep-exv in the real world. We also label the bucket filling volume (if not stuck).}
    \label{fig:real_exv_example}
\end{figure}

11 out of 25 excavation trajectories get stuck and fail to execute successfully for RL-rep-exv. Because the resistive force during rigid objects excavation can be large and only limited amount of force and torque can be applied by the Franka arm in real world. The bucket can get stuck when executing one excavation on the real robot. Examples of robot getting stuck can be seen from Figure~\ref{fig:real_exv_example}. The robot do not get stuck during excavation in simulation, since it can generate relatively large forces. The experimented 2D attacking points for both methods are visualized in Figure~\ref{fig:real_PoA}. The mean and standard deviation of the attack angles are 40.9$^\circ$ and 11.2$^\circ$ for RL-rep-exv, and 53.5$^\circ$ and 15.5$^\circ$ for RL-exv. Similar with simulation, RL-rep-exv attacks further away from the robot base during excavation, which allows it to drag longer and potentially dig more objects. 

In Figure~\ref{fig:real_exv_example}, we qualitatively show 10 physical excavation results of two episodes together with the corresponding excavation volumes. As can be seen, RL-rep-exv is able to plan high-quality excavations for rigid objects. Moreover, the camera pose in the real world totally differs from that in simulation as shown in Figure~\ref{fig:scene_setup}, which demonstrates the camera pose agnostic nature of our point cloud representation.   

\vspace{-5pt}
\begin{table}[ht!]
\centering
    \caption{\label{table:physical_res} Physical experimental results (the bucket filling rate shown in parentheses)}
    \begin{tabular}{
    c@{\hspace{0.35\tabcolsep}}|@{\hspace{0.35\tabcolsep}}
    c@{\hspace{0.35\tabcolsep}}|@{\hspace{0.35\tabcolsep}}
    c@{\hspace{0.35\tabcolsep}}|@{\hspace{0.35\tabcolsep}}
    c@{\hspace{0.35\tabcolsep}}|@{\hspace{0.35\tabcolsep}}
    c@{\hspace{0.35\tabcolsep}}
    c}
    \toprule
      \textbf{Method} & \textbf{Avg. V} & \textbf{Plan R.} & \textbf{Exe. R.} & \textbf{Avg. V w/ Succ. Exe. }\\
      \hline
     RL-rep-exv & 191.0 (42.4\%) & 25/25 & 14/25 & 341.1 (75.8\%) \\
      \hline
     RL-exv & 165.4 (36.8\%) & 17/25 & 15/17 & 275.6 (61.2\%)\\
    \bottomrule
    \end{tabular}
    \label{table:real_exv_res}
\end{table}
\vspace{-5pt}


%% file: 8_discussion.tex
In conclusion, we adopt RL to learn to plan a sequence of attacking poses for rigid objects in clutter.
We learn a novel low-dimensional geometric representation for object point clouds without human labeling, which allows us to perform RL for excavation using a simple policy network. We show the RL algorithm leveraging the learned representation reduces the training time, while achieving similar asymptotic performance compared with a RL algorithm that uses the raw point cloud and a more complex policy network.
Our simulated and physical experiments for cluttered objects excavation show the RL algorithm using the representation outperforms the RL algorithm that uses the raw point cloud.


We directly deploy the RL policy learned from simulation for the real-robot experiments in this work.
As a future work, we plan to extend our geometric representation to mitigate the sim2real visual perception gap by comparing the representation feature distributions between the simulation and the real world.
In the future, we also want to further investigate how to improve the RL performance for excavations with valid plans.
It would be interesting to learn replanning and reactive control for excavation tasks such as mining that involve more perception and physical uncertainty.  
Finally, we would like to adopt the excavation RL from robotic arms to real excavators. 


%% file: 9_appendix.tex





Throughout the experiments, we emulate an excavator with a 7-degrees-of-freedom (DoF) Franka Panda robot arm mounted with a digging bucket, shown in Fig.~\ref{fig:scene_setup}. We only use the shoulder panning, shoulder lifting, elbow lifting , and the wrist lifting joints. We fixate the joint positions of the other 3 joints.  

\subsection{Experiments Setup}
\subsubsection{Simulation Setup}
\label{sec:sim_setup}
We use the PyBullet~\footnote{http://pybullet.org} physics engine for simulation experiments. The simulation scene is shown in Fig.~\ref{fig:sim_scene}, which includes a Franka Panda robot arm mounted with a digging bucket at the end-effector, a digging tray, a dumping tray, and a simulated RGB-D sensor. The robot is tasked with digging the objects in the digging tray and dumping them in the dumping tray. The full volume of the bucket is 450\,cm$^3$. We generate random convex rigid objects by randomly sampling the number of vertices, each a random distance $\in [1$\,cm, $7$\,cm$]$ away from the centroid, and taking the convex hull. We generate a rigid object set with a total of 100k objects and set the density to be 2,700 kg/m$^3$. Scenes are then initialized by drawing from the rigid object set and adding them to the digging tray.

To comply with the robot arm's workspace, we limit the range of the attacking pose $(x, y, \alpha)$: $x \in [-0.37$\,m, $0.29$\,m$]$, $y \in [-0.2$\,m, $0.2$\,m$]$, and $\alpha \in [15^\circ, 120^\circ]$ in the tray frame that is defined in Fig.~\ref{fig:sim_scene}.


\subsubsection{Physical Setup}

    
    
The setup and the tray frame definition for physical experiments is shown in Fig.~\ref{fig:real_scene}. We use the Microsoft Azure RGB-D sensor to capture the point cloud data. We use $282$ rigid wood objects with different geometric shapes and colors for real robot experiments, including $150$ ``Chuckle \& Roar" wood blocks, $100$ ``Melissa \& Doug" wood blocks, and $32$ ``Biubee" wood stone balancing blocks. All the rigid objects are unseen from the RL training. We estimate the density of the wood rigid objects to be $500$ kg$/$m$^3$. We limit the range of the attacking pose $(x, y, \alpha)$: $x \in [-0.37$\,m, $0.15$\,m$]$, $y \in [-0.2$\,m, $0.2$\,m$]$, and $\alpha \in [35^\circ, 100^\circ]$. We make the attacking range in the real world smaller than that in simulation, because the compliant Franka arm could hit joint limits when digging far away from the robot base due to external forces. 

We lay the rigid wood objects on top of red mulch in the digging tray. The relatively deformable red mulch is used as the excavation surface for safety reasons. A small amount of red mulch under the rigid objects can be excavated and dumped sometimes, which we remove when calculating the excavated volume. 

The dumping pose is manually designed and scripted. We use a kitchen scale under the dumping tray to weigh the objects dumped into the tray for each trial. Having the mass of the excavated objects and the objects density, we can compute the volume of excavated objects. 

When the robot penetrates the rigid objects, excessive resistive force could sometimes occur. To alleviate this problem, we command the robot to wiggle the bucket by a small amount during penetration, which we find useful empirically.